%% file: main.tex
\documentclass{article}

\usepackage{booktabs}
\usepackage{graphicx}
\usepackage{microtype}
\usepackage{subfigure}
\usepackage{wrapfig}

\usepackage[pagebackref,breaklinks,colorlinks,bookmarks=true]{hyperref}

\usepackage[accepted]{icml2024}

\usepackage{amsmath}
\usepackage{amssymb}
\usepackage{mathtools}
\usepackage{amsthm}
\usepackage{booktabs}
\usepackage[fixed]{fontawesome5}
\usepackage{multirow}

\usepackage{siunitx}
\renewrobustcmd{\bfseries}{\fontseries{b}\selectfont}
\renewrobustcmd{\boldmath}{}

\theoremstyle{plain}

\theoremstyle{definition}

\theoremstyle{remark}

\definecolor[named]{xBlue}{HTML}{18647E}
\definecolor[named]{xOrange}{HTML}{FF9B00}
\definecolor[named]{xGray}{HTML}{808080}
\definecolor[named]{xGreen}{HTML}{60B950}
\definecolor[named]{xRed}{HTML}{A30B37}
\definecolor[named]{xDarkBlue}{cmyk}{1,0.58,0,0.21}

\definecolor[named]{cPink}{HTML}{F26DF9}

\definecolor[named]{TLDRViolet}{HTML}{800080}
\definecolor[named]{MissingCyan}{HTML}{47D4FF}
\definecolor[named]{PendingSaffron}{HTML}{F26DF9}
\definecolor[named]{SkeletonGray}{HTML}{808080}

\definecolor[named]{SiddBlue}{HTML}{18647E}
\definecolor[named]{SurajPink}{HTML}{F26DF9}
\definecolor[named]{AshwinYellow}{HTML}{F0C808}
\definecolor[named]{PercyRed}{HTML}{A30B37}
\definecolor[named]{TomGreen}{HTML}{21D19F}
\definecolor[named]{DorsaOrange}{HTML}{FF9B00}

\hypersetup{
    colorlinks=true,
    linkcolor=xOrange,
    filecolor=magenta,      
    urlcolor=xDarkBlue,
    citecolor=xBlue,
}

\usepackage{epigraph}
\usepackage{inconsolata}

\def\Snospace~{\S{}}

\def \papermode{draft}  

\def \draftmode{draft}

\ifx \papermode \draftmode
    
    \newcommand{\needcite}{{\color{MissingCyan}[NEED CITE]}}
    \newcommand{\tldr}[1]{{\color{TLDRViolet}{[TL;DR] :: \textit{#1}}}}

    \newcommand{\makecomment}[3]{{\color{#2}[\textbf{#1}]: #3}}

\else
    
    \newcommand{\needcite}[1]{}
    \newcommand{\warning}[1]{}
    \newcommand{\tldr}[1]{}
    \newcommand{\maybe}[1]{}
    \newcommand{\makecomment}[3]{}
\fi

\icmltitlerunning{Investigating the Design Space of Visually-Conditioned Language Models}

\begin{document}

\twocolumn[
\icmltitle{Prismatic VLMs: Investigating the Design Space of \texorpdfstring{\\}{} Visually-Conditioned Language Models}



\icmlsetsymbol{equaladv}{\textdagger}

\begin{icmlauthorlist}
\icmlauthor{Siddharth Karamcheti}{stanford,tri}
\icmlauthor{Suraj Nair}{tri}
\icmlauthor{Ashwin Balakrishna}{tri}
\icmlauthor{Percy Liang}{stanford}
\icmlauthor{Thomas Kollar}{tri,equaladv}
\icmlauthor{Dorsa Sadigh}{stanford,equaladv}

\medskip
\begin{tabular}{c@{\hskip 21pt}c}
    \raisebox{-1pt}{\faGithub} \href{https://github.com/TRI-ML/prismatic-vlms}{\small \path{github.com/TRI-ML/prismatic-vlms}} & 
    \raisebox{-1.5pt}{\faChartBar[regular]} \href{https://github.com/TRI-ML/vlm-evaluation}{\small \path{github.com/TRI-ML/vlm-evaluation}} \\
\end{tabular}
\end{icmlauthorlist}

\icmlaffiliation{stanford}{Department of Computer Science, Stanford University, Stanford, CA, USA}
\icmlaffiliation{tri}{Toyota Research Institute, Los Altos, CA, USA} 
\icmlcorrespondingauthor{Siddharth Karamcheti}{\url{skaramcheti@cs.stanford.edu}}

\icmlkeywords{Visually-Conditioned Language Models, Multimodal Pretraining, Large Language Models}

\vskip 0.3in
]



\printAffiliationsAndNotice{\icmlEqualAdvising} 

\begin{abstract}
\input{sections/00_abstract}
\end{abstract}

\section{Introduction}
\label{sec:introduction}
\input{sections/01_introduction}

\newpage

\section{Preliminaries}
\label{sec:preliminaries}
\input{sections/02_preliminaries}

\section{Evaluation Suite}
\label{sec:evaluation}
\input{sections/03_evaluation}

\section{Experiments -- Investigating Design Axes}
\label{sec:experiments}
\input{sections/04_experiments}

\subsection{Optimization Procedure}
\label{subsec:optimization-procedure}
\input{sections/04a_optimization-procedure}

\subsection{Image Processing \& Visual Representations}
\label{subsec:visual-representations}
\input{sections/04b_visual-representations}

\subsection{Integrating Language Models}
\label{subsec:language-models}
\input{sections/04c_language-models}

\subsection{Scaling Properties: Training Time \& Data}
\label{subsec:scaling-properties}
\input{sections/04d_scaling-properties}

\section{\textsc{Prism} -- Distilling Key Insights}
\label{sec:prism-insights}
\input{sections/05_prism-insights}

\section{Limitations \& Future Work}
\label{sec:limitations-future-work}
\input{sections/06_limitations-future-work}

\section{Conclusion}
\label{sec:conclusion}
\input{sections/07_conclusion}


\clearpage
\section*{Impact Statement}
\label{sec:broader-impacts}
\input{sections/xA_broader-impacts}

\section*{Acknowledgements}
Toyota Research Institute (``TRI'') and the Office of Naval Research (ONR \#N00014-22-1-2293) provided funds to support this work. Siddharth Karamcheti is supported by the Open Philanthropy Project AI Fellowship. Finally, we thank Adrien Gaidon, Rares Ambrus, Achal Dave, Blake Wulfe, Masha Itkina, Jean Mercat, Igor Vasiljevic, Sedrick Keh, Kushal Arora, John Thickstun, and David Hall for their insight and advice during the development of this work.

\newpage
\bibliography{x-refdb}
\bibliographystyle{icml2024}

\newpage
\appendix
\onecolumn

\input{sections-appendix/x0_additional-figures}

\clearpage

\section{Training Visually-Conditioned Language Models}
\label{appx:implementation-training}
\input{sections-appendix/xA_implementation-training}

\newpage
\section{Evaluation Protocol}
\label{appx:evaluation-protocol}
\input{sections-appendix/xB_evaluation-protocol}

\clearpage
\input{sections-appendix/xx_tabulated-results}


\end{document}

%% file: sections/00_abstract.tex

Visually-conditioned language models (VLMs) have seen growing adoption in applications such as visual dialogue, scene understanding, and robotic task planning; adoption that has fueled a wealth of new models such as LLaVa, InstructBLIP, and PaLI-3. Despite the volume of new releases, key design decisions around image preprocessing, architecture, and optimization are under-explored, making it challenging to understand what factors account for model performance -- a challenge further complicated by the lack of objective, consistent evaluations. To address these gaps, we first compile a suite of standardized evaluations spanning visual question answering, object localization, and challenge sets that probe properties such as hallucination; evaluations that provide fine-grained insight VLM capabilities. Second, we rigorously investigate VLMs along key design axes, including pretrained visual representations and training from base vs. instruct-tuned language models, amongst others. We couple our analysis with three resource contributions: (1) a unified framework for evaluating VLMs, (2) optimized, flexible training code, and (3) checkpoints for all models, including a family of VLMs at the 7-13B scale that strictly outperform InstructBLIP and LLaVa v1.5, the state-of-the-art in open VLMs.

%% file: sections/01_introduction.tex

\begin{figure}[t!]
    \vspace*{-2mm}
    \centering
    \includegraphics[width=\linewidth]{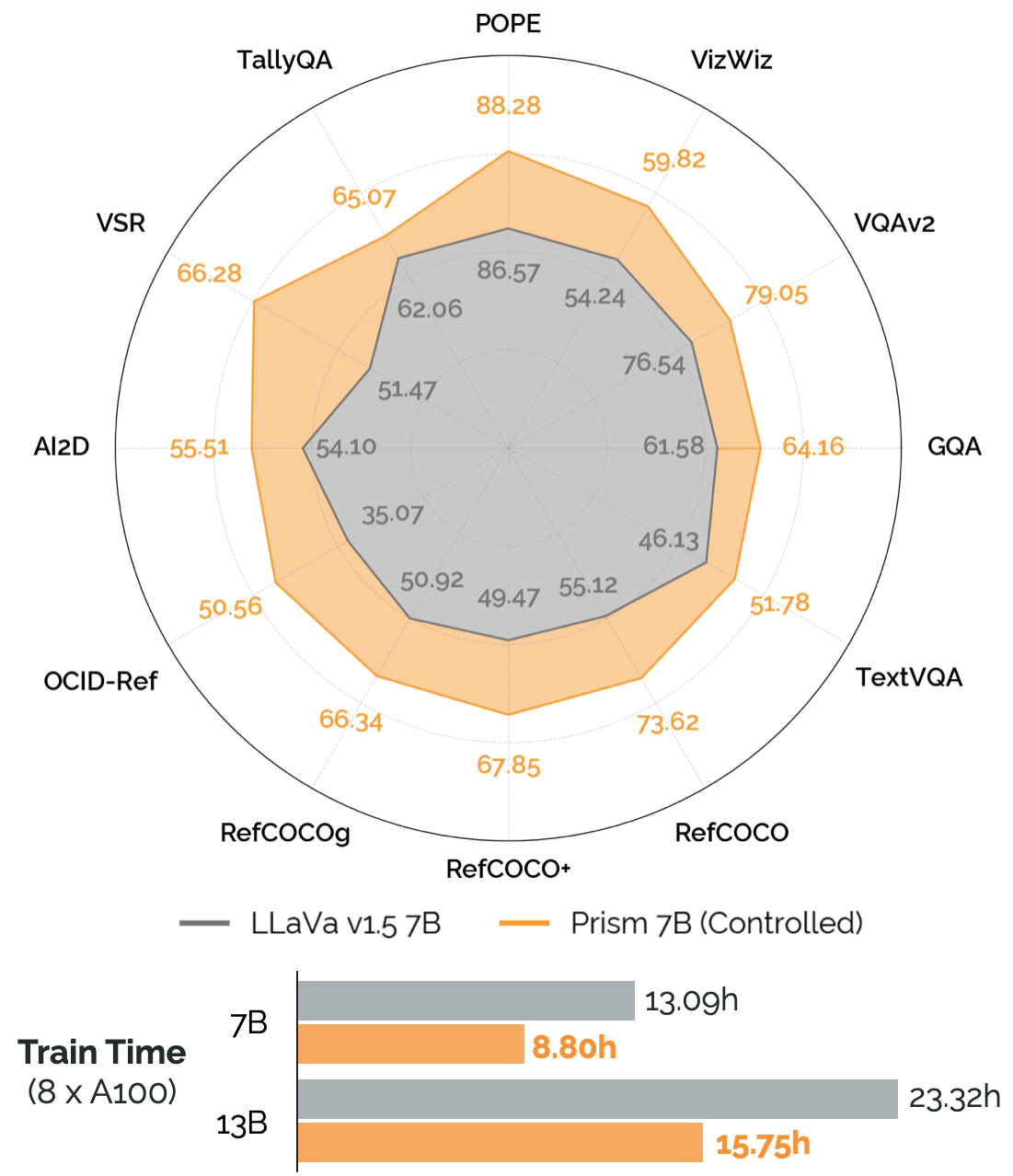}
    \vspace*{-7mm}
    \caption{\textbf{Prismatic VLMs.}{\renewcommand*{\thefootnote}{\fnsymbol{footnote}}\setcounter{footnote}{0}\footnotemark{}}
    Through rigorous experiments exploring the design space of visually-conditioned language models (VLMs), we identify insights that improve training. When controlling for data and scale, our models ({\color{xOrange} orange}) outperform the state-of-the-art LLaVa v1.5 \citep[{\color{xGray} gray;}][]{liu2023llavav15} across 12 diverse tasks, \textit{while saving more than 30\% the training compute}.}
    \label{fig:results-overview}
    \vspace*{-5mm}
\end{figure}

\begin{figure*}[t]
    \vspace*{-2mm}
    \centering
    \includegraphics[width=\textwidth]{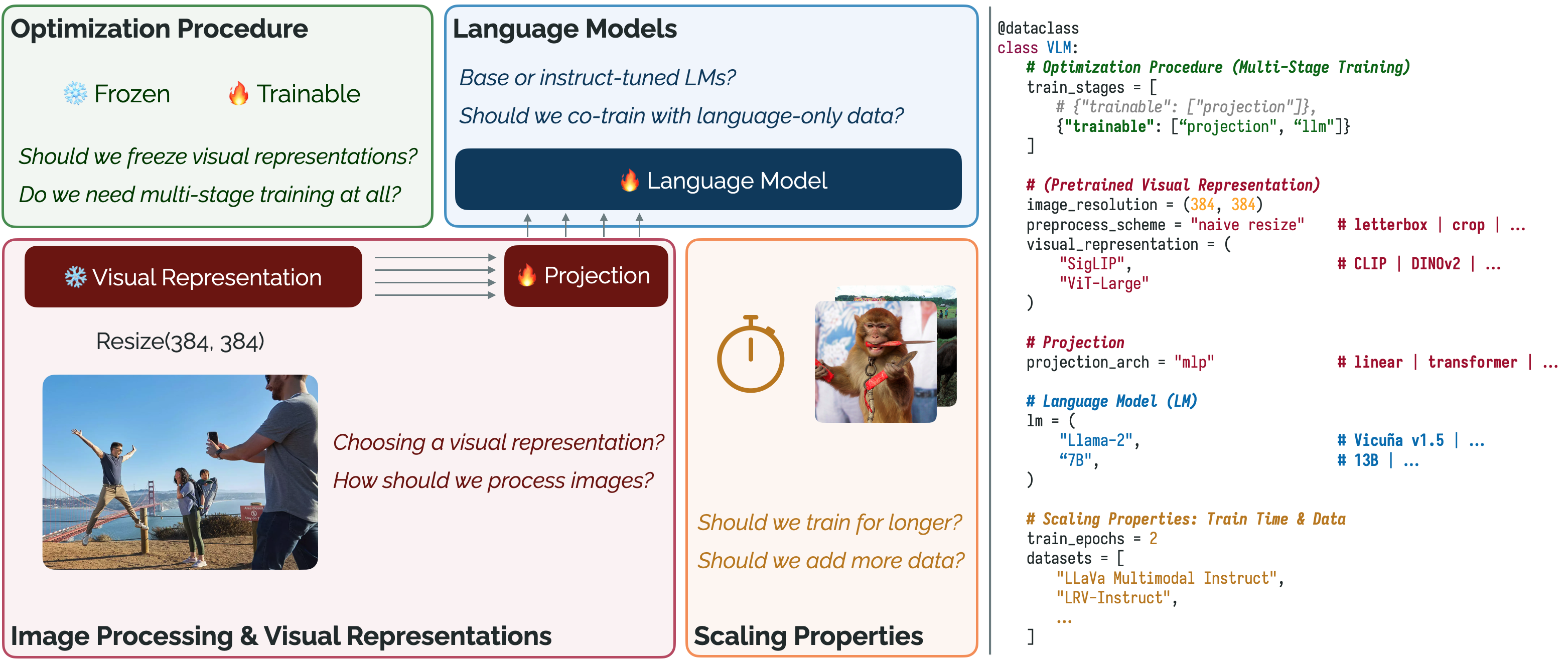}
    \vspace*{-7mm}
    \caption{\textbf{Exploring VLM Design Axes.} We explore four key design axes for developing VLMs: 1) optimization procedure, 2) image processing and pretrained visual representations, 3) language models, and 4) scaling properties around training time and data (\textbf{left}). To enable this exploration, we make a key resource contribution: an open-source, flexible codebase for efficiently training VLMs (\textbf{right}).}
    \label{fig:design-axes}
    \vspace*{-3mm}
\end{figure*}

\vspace*{-1.21mm}
\epigraph{
    \textit{If you have built castles in the air, your work need not be lost; that is where they should be. \textbf{Now put the foundations under them.}}
}{
    \textsc{--- Henry David Thoreau}
}
\vspace*{-2.1mm}

\newpage

{\renewcommand*{\thefootnote}{\fnsymbol{footnote}}\footnotetext{\textbf{Prismatic} (adj) -- \textit{relating to or having the form of a prism}. \\
Like a geometric prism, our VLMs share a common structure, but are characterized by different ``faces'' -- the individual design axes we explore in this work.}\setcounter{footnote}{0}}


Visually-conditioned language models (VLMs) generate natural language responses from image input and text prompts, providing a general, expressive interface for a growing spectrum of applications -- grounded chat \citep{li2023videochat, gong2023multimodalgpt}, visual programming \citep{suris2023vipergpt, subramanian2023modularvqa}, robotic control \citep{driess2023palme, brohan2023rt2}, etc. This broad adoption is fueled by a recent \textit{paradigm shift} in how we develop VLMs; eschewing the complex architectures and training objectives of prior work \citep{tan2019lxmert, li2022blip, li2023blip2}, new VLMs adopt a simple approach, treating patch features from pretrained visual backbones \citep[e.g., CLIP;][]{radford2021clip} as tokens that can be projected into the input space of a language model (LM). This ``patch-as-token'' approach enables training with a simple objective -- next-token prediction -- and allows us to harness the ecosystem of powerful LMs such as Llama-2 and Mistral \citep{touvron2023llama2, jiang2023mistral}, along with the tools to efficiently train them \citep[e.g., FSDP;][]{zhao2023fsdp}. This combination has fueled the rapid development and release of models such as LLaVa v1.5, and PALI-3 that adopt the same underlying recipe, while varying individual ingredients such as the choice of pretrained components, data, or optimization procedure \citep{liu2023llavav15, chen2023pali3}.

Unfortunately, existing approaches only cover a sliver of the design space around building and training VLMs, without thoroughly evaluating the impact of given choices on downstream capabilities. This motivates the key question of this work: \textit{what are the key design decisions that influence VLM capabilities and downstream use}? To provide answers to this question, we first need a way to \textbf{thoroughly evaluate} the strengths and weaknesses of a given model. Doing this effectively requires compiling a standardized evaluation suite comprised of tasks that are diverse and objective; crucially, these tasks should allow for probing specific capabilities such as spatial reasoning, out-of-distribution generalization, and commonsense understanding, amongst others. Second, we need to \textbf{rigorously explore} different VLM design axes, not only to build a concrete set of recommendations, but to tie individual choices to downstream performance.

This work addresses these axes through four contributions. First, to provide fine-grained insight into VLM capabilities, \textbf{we compile a standardized evaluation suite} comprised of twelve benchmarks from the vision-and-language literature, including four tasks spanning visual question answering \citep{bigham2010vizwiz, goyal2017making, hudson2019gqa, singh2019textvqa}, four tasks spanning object localization \citep{kazemzadeh2014refcoco, yu2016refcoco, wang2021ocidref}, and four challenge tasks evaluating fine-grained spatial reasoning, hallucination, and diagram understanding \citep{acharya2018tallyqa, liu2022vsr, li2023pope, kembhavi2016ai2d}. Second, \textbf{we develop an optimized and modular codebase for VLM training} that emphasizes flexibility, allowing users to easily swap in pretrained components, optimization procedures, data, and more (\autoref{fig:design-axes}; right). Third, we use these resource contributions to perform \textbf{targeted experiments exploring four key design axes} (\autoref{fig:design-axes}; left): 1) optimization procedure, 2) image processing and visual representations, 3) language models, and 4) scaling training time and data. We identify a number of insights; for example, we find that multi-stage training procedures adopted by existing work can be eliminated without impact on performance, reducing compute costs by 20-25\%. We also find that \emph{fused} visual backbones that merge features from different backbones such as CLIP \citep{radford2021clip} and DINOv2 \cite{oquab2023dinov2} lead to more performant VLMs across the board. Finally, we consolidate our findings and train a family of models -- \textsc{Prism}s -- at the 7B/13B scale that \textbf{strictly outperform state-of-the-art open VLMs} such as InstructBLIP and LLaVa v1.5.\footnote{
We release our optimized training codebase, evaluation suite, \\
\hspace*{14pt} and \textit{checkpoints for all models trained as part of this work}. \\[3pt]
\hspace*{21pt}\raisebox{-0.5pt}{\faGithub} \href{https://github.com/TRI-ML/prismatic-vlms}{\path{github.com/TRI-ML/prismatic-vlms}} \\[1.5pt]
\hspace*{21pt}\raisebox{-1pt}{\faChartBar[regular]} \href{https://github.com/TRI-ML/vlm-evaluation}{\path{github.com/TRI-ML/vlm-evaluation}}
}

%% file: sections/02_preliminaries.tex

\begin{figure*}[t]
    \vspace*{-2mm}
    \centering
    \includegraphics[width=\textwidth]{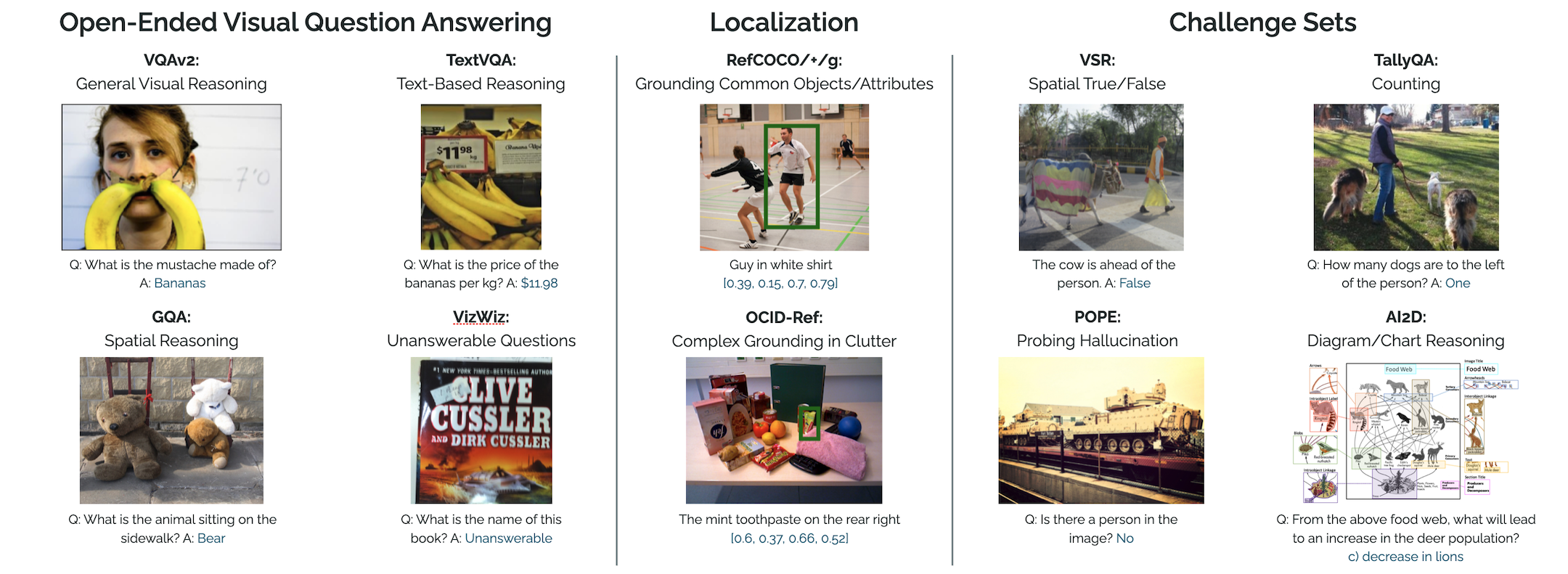}
    \vspace*{-9mm}
    \caption{\textbf{Evaluation Suite Overview.} We compile multiple established benchmarks spanning visual question answering, localization, and challenge tasks (e.g., evaluating counting, spatial relationships, propensity to hallucinate). This evaluation suite forms the backbone for all of our analysis, giving us fine-grained insight into the impact of individual VLM design choices.}
    \label{fig:evaluation-suite}
    \vspace*{-3mm}
\end{figure*}

To ground our analysis, we require 1) a VLM model architecture, 2) pretraining data, and 3) a training implementation. 

\noindent \textbf{Model Architecture.} We adopt the general architecture used by many recent VLMs, such as LLaVa, Qwen-VL, and PaLI-3 \citep{liu2023llava, bai2023qwenvl, chen2023pali3}. These architectures use a (pretrained) visual backbone to map an input image to a sequence of patch features that are then projected individually into the embedding space of an LM. Formally, a VLM takes as input an image $x_\text{img} \in \mathbb{R}^{H \times W \time 3}$ and text prompt tokens $u_\text{prompt}$ with arbitrary sequence length $K$. These inputs are then fed to the following components: 1) a visual representation backbone, 2) a vision-language projector, and 3) a language model.

\noindent \textit{Visual Representation}. We first process $x_\text{img}$ subject to a visual representation backbone $V_\omega$ that outputs a sequence of features $p_\text{img} \in \mathbb{R}^{L \times h_{\text{vision}}}$ where $p_\text{img} = V_\omega(x_\text{img})$. As an example, $p_{\text{img}}$ might be the patch features output by a Vision Transformer \citep[ViT;][]{dosovitskiy2021vit}. 

\noindent \textit{Vision-Language Projector}. Next, we map $p_\text{img}$ to a sequence of \textit{embeddings} $e_\text{img} \in \mathbb{R}^{L \times h_{\text{text}}}$ via a learned projector $F_\psi$, where $e_\text{img} = F_\psi(p_\text{img})$.

\noindent \textit{Language Model}. Finally, we concatenate the sequence $e_\text{img}$ with the text prompt embeddings $e_\text{prompt} = \text{embed}(u_\text{prompt})$, passing the result to the language model. The language model generates output text $u_\text{gen} = \text{LM}_\theta([e_\text{img}; e_\text{prompt}])$.

The composition $\text{LM}_\theta([F_\psi(V_\omega(o_\text{rgb})); \text{embed}(u_\text{prompt})])$ then defines a VLM. Given a triple $(x_\text{img}, u_\text{prompt},  \hat{u}_\text{gen})$ during training, we minimize the loss $\mathcal{L}(\omega, \psi, \theta) = - \log p(\hat{u}_\text{gen} \mid x_\text{img}, u_\text{prompt})$ via gradient descent.

\noindent \textbf{Pretraining Dataset.} We limit our selection of pretraining data to datasets that are fully open-source (e.g., under permissive research licenses), and that have been used in prior work. Specifically, we use the LLaVa v1.5 data mixture, which consists of two subsets used for a multi-stage training pipeline. The first subset consists of a 558K sample mixture of examples sourced from various captioning datasets \citep[e.g., Conceptual Captions, LAION][]{sharma2018conceptual, schuhmann2021laion400m}, while the second consists of 665K multimodal instruct tuning examples comprised of synthetic data generated in \citet{liu2023llava}, as well as examples from existing vision-language training sets \citep[e.g., GQA, TextCaps;][]{hudson2019gqa, sidorov2020textcaps}, and notably, a sample of language-only data from ShareGPT \citep{sharegpt2023sharegpt}. We provide a comprehensive breakdown of the pretraining data mixture in \autoref{appx-subsec:pretraining-data}.

\noindent \textbf{Training Implementation \& Verification.} To investigate the design axes enumerated in \autoref{sec:introduction}, we require code for VLM training that is \textit{efficient} and \textit{flexible}; critically, we need the ability to easily swap out vision and LM backbones and handle arbitrary optimization procedures (e.g., freezing the vision backbone during training). With these requirements, we implement our training codebase in PyTorch, using Fully Sharded Data Parallel \citep[FSDP;][]{zhao2023fsdp} and BF16 mixed precision. FSDP lets us specify precision for individual model components (e.g., FP16 for vision backbones, BF16 for LMs), enables portability to different hardware, and provides minimal implementation overhead. Following reproducibility practices from prior work \citep{karamcheti2021mistral, biderman2023pythia}, we fix initialization randomness and fix batch order during training. We leverage TIMM \citep{wightman2019timm} and Hugging Face Transformers \citep{wolf2019transformers} to provide pretrained models.

To validate our code, we run an apples-to-apples reproduction of LLaVa v1.5 \citep{liu2023llavav15} at both the 7B and 13B parameter scale. Successful reproduction results are in \autoref{fig:reproduction-pipeline} (left). We find our implementation is considerably more efficient than the reference LLaVa v1.5 training implementation: when benchmarked on the same hardware (an AWS \texttt{p4de.24xlarge} node with 8 A100 GPUs), we observe 20\% faster step times with our FSDP-backed implementation, a notable gain given LLaVa leverages the well-optimized DeepSpeed ZeRO library \citep{rasley2020deepspeed}.

\textbf{We highlight this open-source training codebase as one of the key contributions of this work}. Unlike other open codebases, we provide a modular and expressive interface for easily specifying or adding model components, optimization procedures, and data with minimal code changes (\autoref{fig:design-axes}; right). In providing an efficient and easily extensible framework, we enable future research around designing new evaluations, developing and training new VLMs, and finetuning or otherwise adapting existing models for diverse downstream applications -- all while maintaining a high standard of reproducibility and controlled experimentation.

%% file: sections/03_evaluation.tex

The first contribution of this work is a unified evaluation suite that offers \textit{fine-grained insight} into the capabilities of a given VLM. Recent work in evaluating VLMs tends to rely on automated evaluations that use powerful LMs such as GPT-4 \citep{openai2023gpt4} to judge relative and subjective performance\citep{liu2023mmbench, yu2023mmvet}, making it hard to measure the absolute impact of a given design change. Instead, we focus on evaluations with well-defined metrics, spanning the following three areas:

\noindent \textit{Open-Ended Visual Question Answering.} We evaluate on VizWiz \citep{bigham2010vizwiz}, VQAv2 \citep{goyal2017making}, GQA \citep{hudson2019gqa}, and TextVQA \citep{singh2019textvqa}. Both VizWiz and VQAv2 assess general visual reasoning; VizWiz also contains a series of unanswerable questions. GQA evaluates spatial reasoning, while TextVQA assesses reasoning around text (e.g., labels, signage) present in an image.

\noindent \textit{Localization.} Part of the pretraining data mixture (from \autoref{sec:preliminaries}) contains examples of predicting normalized bounding box coordinates given referring expressions in language. As such, we evaluate bounding box prediction accuracy on RefCOCO, RefCOCO+, and RefCOCOg \citep{kazemzadeh2014refcoco, yu2016refcoco}, and on OCID-Ref \citep{wang2021ocidref}. RefCOCO focuses on short descriptions with spatial anchors, RefCOCO+ on strictly appearance based descriptions, and RefCOCOg on long, rich descriptions; OCID-Ref is a robotics dataset probing out-of-distribution generalization, with a focus on localizing objects in clutter.

\noindent \textit{Challenge Sets (Closed-Set Prediction).} We evaluate on Visual Spatial Reasoning \citep[VSR;][]{liu2022vsr}, TallyQA \citep{acharya2018tallyqa}, POPE \citep{li2023pope}, and AI2 Diagrams \citep[AI2D;][]{kembhavi2016ai2d}. VSR consists of challenging True/False questions about individual spatial relationships in diverse scenes (e.g., ``the cake is at the edge of the dining table''); this is an especially challenging task, with most existing models failing to outperform the majority class baseline (51\%). TallyQA consists of questions that assess a VLM's ability to count objects described in language, with expressions that range in complexity. POPE consists of targeted Yes/No questions that assess a VLM's propensity to hallucinate. Finally, AI2D consists of multiple choice questions about scientific diagrams and charts, many of which require reading labels or text annotations (e.g., flowchart labels, plot axes).

We use the validation sets for all benchmarks except GQA (where use the recommended the test-dev split), VSR (where we use the zero-shot test split), and POPE (where there is only a single evaluation split). We provide further detail around evaluation protocols in \autoref{appx:evaluation-protocol}.

%% file: sections/04_experiments.tex

\begin{figure*}[t]
    \vspace*{-2mm}
    \centering
    \includegraphics[width=\textwidth]{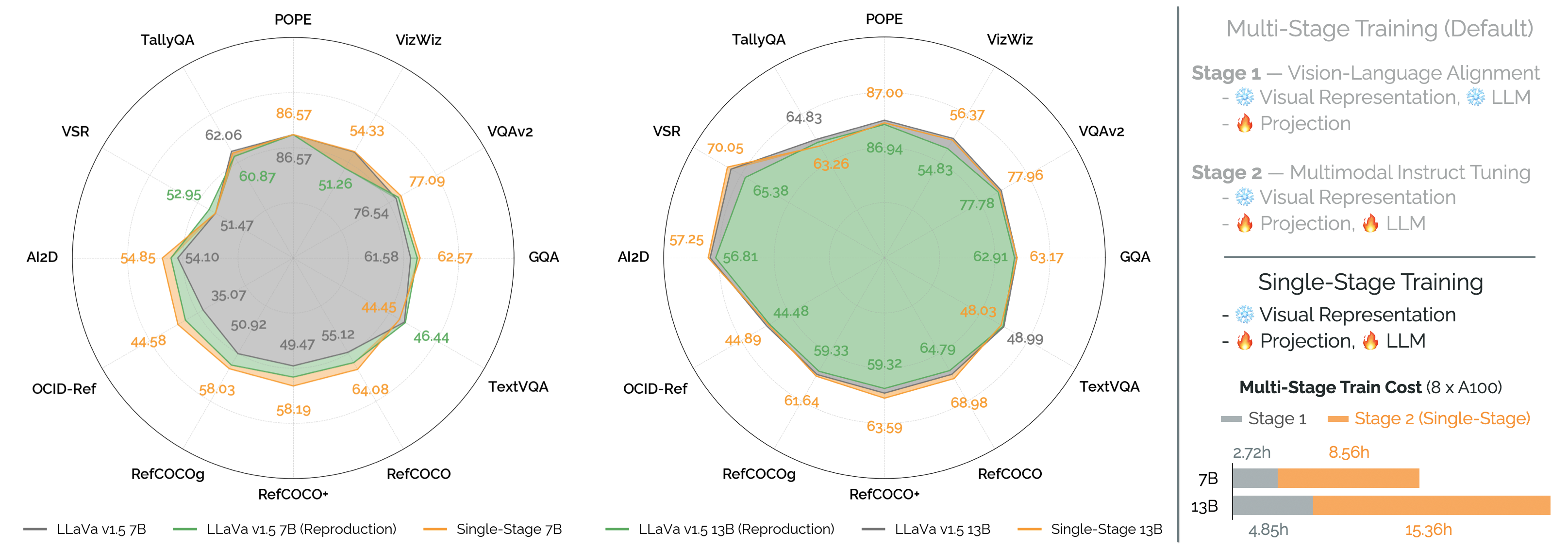}
    \vspace*{-7mm}
    \caption{\textbf{Reproducing LLaVa v1.5 \& Exploring Optimization Procedures.} To validate our training codebase (\autoref{sec:preliminaries}), we reproduce LLaVa v1.5 ({\color{xGreen} green}), with our models reproducing the performance reported in \citet{liu2023llavav15} ({\color{xGray} gray}). We then run our first experiment (\autoref{subsec:optimization-procedure}) investigating the need for expensive multi-stage training (\textbf{right}). We find that single-stage training produces VLMs \textit{that maintain or outperform multi-stage models} ({\color{xOrange} orange}), saving considerable compute; as a result, we carry this change forward to all future experiments.}
    \label{fig:reproduction-pipeline}
    \vspace*{-3mm}
\end{figure*}

Our second contribution is a series of targeted experiments exploring the VLM design space along \textit{four key axes}: (\autoref{subsec:optimization-procedure}) optimization procedure, (\autoref{subsec:visual-representations}) image processing and visual representations, (\autoref{subsec:language-models}) language models, and (\autoref{subsec:scaling-properties}) scaling properties such as training time and data diversity. 

\noindent \textbf{Experiment Design: Protocols \& Drawing Conclusions.} We first validate our VLM training implementation by reproducing LLaVa v1.5 (see \autoref{sec:preliminaries}), adopting the design choices of the original work -- the same choices used by many other recent VLMs: ``letterbox padding'' to process images, CLIP ViT-Large with a patch size of 14 and input resolution of 336px \citep[CLIP ViT-L/14 @ 336px;][]{radford2021clip} as the visual representation, Vicuña v1.5 as the LM backbone, and the two-stage training pipeline using both data subsets described in \autoref{sec:preliminaries}. Successful reproduction results at both the 7B and 13B scale are in \autoref{fig:reproduction-pipeline} (left). Given both the fidelity of our reproduction and the prevalence of these design choices, we anchor our analyses around this parameterization. Critically, each of the experiments in \autoref{subsec:visual-representations}, \autoref{subsec:language-models}, and \autoref{subsec:scaling-properties} are formulated as \textit{single-step changes of this base architecture, with all other choices held constant}.

As each evaluation in \autoref{sec:evaluation} uses different metrics with different scales, direct comparison is challenging. We address this by computing normalized Z-scores for each model and evaluation (using the mean and standard deviation across all models). These scores are used to compute statistical significance (further details in \autoref{appx-subsec:evaluation-comparison-significance}), and to set the relative scales of each radar plot (for completeness, we also provide the absolute metrics as colored and bolded labels).

%% file: sections/04a_optimization-procedure.tex

\begin{figure}[t]
    \centering
    \includegraphics[width=0.85\linewidth]{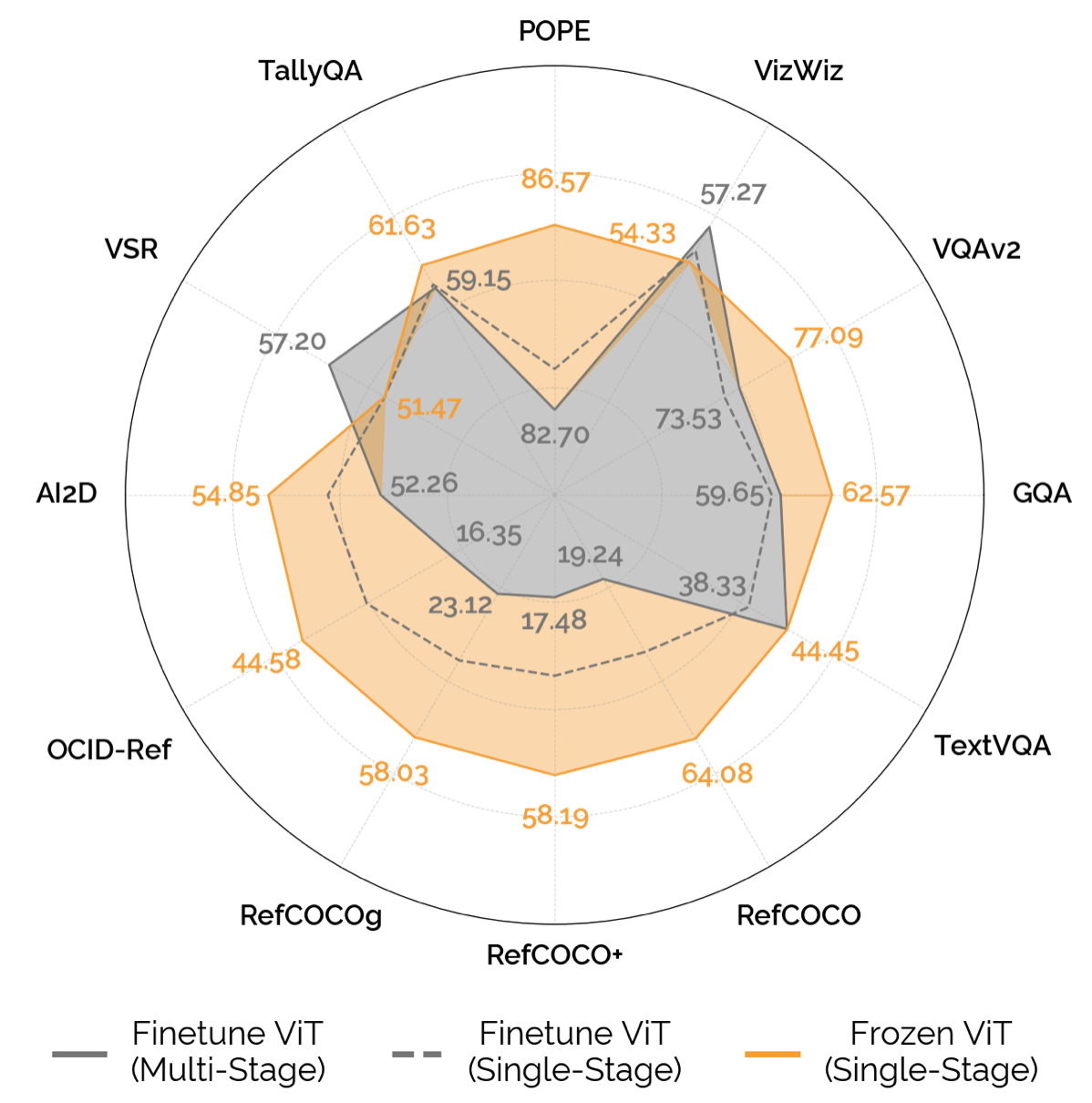}
    \vspace*{-3mm}
    \caption{\textbf{Full Finetuning through Visual Backbones.} We explore the impact of finetuning the (conventionally frozen) visual backbone in addition to the projector and language model during training. We see that in both the single and multi-stage paradigms, finetuning the vision backbone dramatically degrades performance across almost all benchmarks -- especially on localization tasks.}
    \label{fig:finetune-vision-backbone}
    \vspace*{-5mm}
\end{figure}

\begin{figure*}[t!]
    \vspace*{-2mm}
    \centering
    \includegraphics[width=\textwidth]{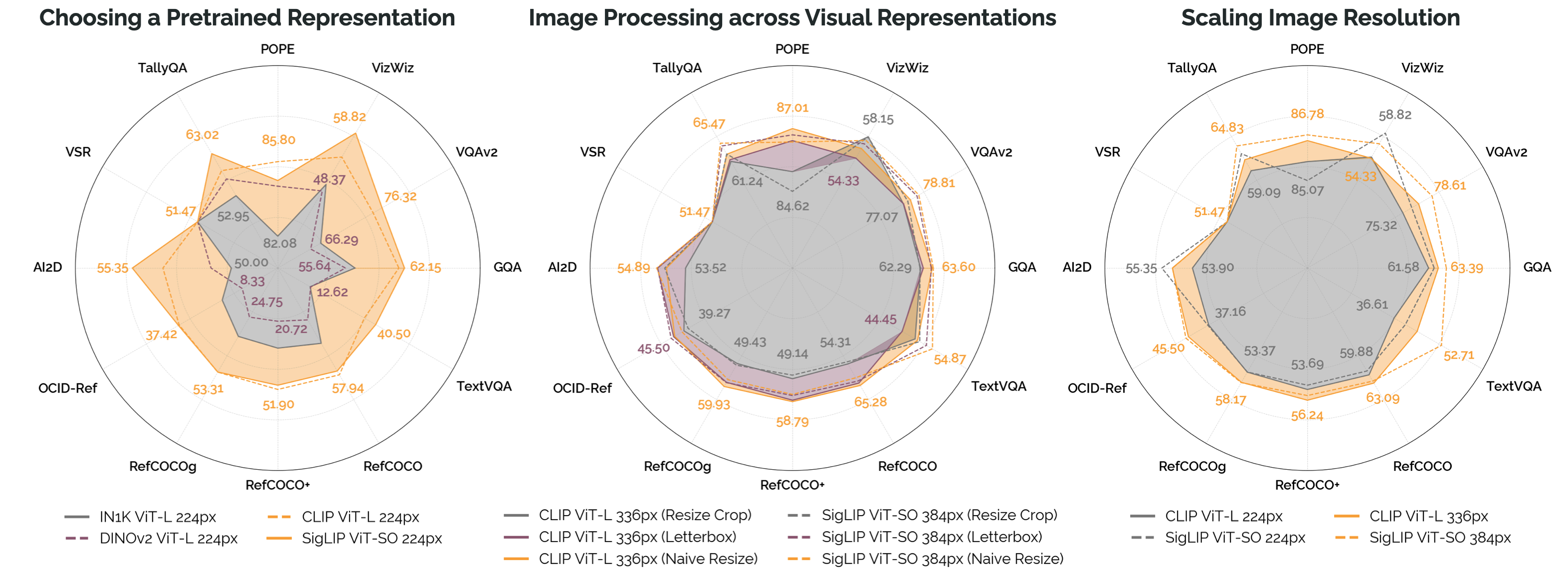}
    \vspace*{-7mm}
    \caption{\textbf{Image Processing \& Visual Representations}. We explore choices around image processing and visual representations in \autoref{subsec:visual-representations}. Specifically, we investigate the impact of different visual representations (\textbf{left}), how performance varies as function of image processing strategies (\textbf{middle}), and the impact of increasing input image resolution (\textbf{right}).}
    \label{fig:visual-rep-results}
    \vspace*{-3mm}
\end{figure*}

In this section we focus on design choices around the optimization procedure used to initialize and train each of the three components described in \autoref{sec:preliminaries}. Specifically, we examine the effects of multi-stage training where different VLM components are frozen at different points in training.

\noindent{\textbf{Multi-Stage Training.}} One of the prevalent design choices adopted by many VLMs \citep{chen2023shikra, ye2023mplugowl} is the inclusion of a two-stage training pipeline: (1) an alignment stage to align vision and language features by training the randomly initialized projector $F_\psi$ in isolation, freezing all other components (\autoref{fig:reproduction-pipeline}, right) and (2) a finetuning stage, where only the visual representation is frozen while both the projection and LM are trained.

Adopting multi-stage training complicates implementation and adds to training cost; therefore, as an initial experiment, we evaluate the need for this first stage through a targeted ablation. We compare the default two-stage training procedure with a \textit{single-stage} approach that skips directly to finetuning $F_\psi$ and the LM. We find (\autoref{fig:reproduction-pipeline}; left) that including the explicit projector pretraining stage is unnecessary, with single-stage training improving aggregate performance ($p = \text{0.00558}$). Eliminating this first stage saves 20-25\% of training cost, and removes the need for additional, stage-specific data (e.g., the captioning subset from \autoref{sec:preliminaries}). \textit{As this change strictly improves performance and efficiency, we adopt single-stage training for all following experiments.}

\noindent{\textbf{Full Finetuning through Visual Backbones.}} Another popular design choice in existing VLMs that leverage pretrained visual representations is to leave the visual backbone \textit{frozen} during the entirety of training \citep{liu2023llavav15, driess2023palme, li2023blip2}. Such a choice limits the potential to learn improved visual representations conducive to language generation during the course of training. Thus, we ask -- \textit{is there potential to improve VLM performance by finetuning the full model, including the visual backbone?} We find (\autoref{fig:finetune-vision-backbone}) that this is not the case, and that finetuning the visual backbone significantly degrades performance ($p = \text{0.00381}$), especially on tasks requiring fine-grained spatial reasoning such as RefCOCO and OCID-Ref. 

\noindent \textit{Remark}. The degraded performance from full finetuning could be for a number of reasons ranging from the scale and diversity of the vision-language data we train on to language generation as a learning objective (vs. objectives that encourage learning fine-grained perceptual features). Especially given the existence of closed-source models such as Fuyu-8B \citep{adeptai2023fuyu} that adopt this paradigm to great success, we believe that identifying ways to prevent such feature collapse during VLM training (e.g., via auxiliary objectives) to be a rich direction for future work.

%% file: sections/04b_visual-representations.tex


\noindent \textbf{Choosing a Pretrained Vision Representation.} CLIP \citep{radford2021clip} has become the default choice for visual representation for almost all VLMs, despite a wealth of visual representations trained on diverse data sources. In this experiment, we perform a head-to-head comparison between CLIP, SigLIP \citep{zhai2023siglip}, DINOv2 \citep{oquab2023dinov2}, and a standard Vision Transformer pretrained for classification \citep[on ImageNet-21K, finetuned on ImageNet-1K;][]{dosovitskiy2021vit, steiner2021vitaug}; for fair comparison, we use the ViT-Large model variant.\footnote{To evaluate on an image resolution common to all representations (224px), we use the \textit{shape-optimized} SigLIP model \citep[ViT-SO][]{alabdulmohsin2023vitso} that is slightly larger than a ViT-Large at 400M parameters (vs 307M).} We find (\autoref{fig:visual-rep-results}; left) that the backbones trained with vision-language contrastive objectives (i.e., CLIP, SigLIP) are significantly more performant than alternatives ($p = \text{7.11e-8}$). 

\noindent \textit{Remark}. While the vision-language contrastive objective is one explanation for the strengths of CLIP and SigLIP, another possible explanation is one of training image distribution. Both CLIP and SigLIP contain internet-sourced images (e.g., sketches, diagrams, animated graphics, etc.) not in ImageNet or in the DINOv2 pretraining data.

\noindent \textbf{Image Processing across Visual Backbones.} Most images have resolutions and aspect ratios that widely vary, yet most visual backbones expect square images at a fixed size; to reconcile this, the overwhelming default is to ``resize \& crop'' an image to size. While this tends to work well for applications such as classification, cropping out parts of an image is especially harmful for tasks requiring full-scene reasoning. In this experiment, we evaluate three different image processing schemes -- the default ``resize \& crop'' scheme, the ``letterbox padding'' scheme used by LLaVa v1.5 that pads non-square images to square, and a ``naive resize'' scheme that warps the original image aspect ratio, squeezing or stretching an image to square. Our findings (\autoref{fig:visual-rep-results}; middle) are surprising: while cropping is clearly suboptimal, the ``naive resize'' scheme is the most performant for CLIP. For SigLIP, both ``naive resize'' and ``letterbox padding'' perform similarly. In general, our results favor ``naive resizing'' over ``letterbox padding'' but we cannot rule the improvement statistically significant ($p = \text{0.0176}$).

\noindent \textit{Remark}. Two speculative arguments for naively resizing an image over padding are those of minimizing ``dead pixels'' and distribution shift. An image with a 16:9 aspect ratio that is padded to square introduces a large amount of uninformative pixels (exceeding 40\%); warping the aspect ratio is possibly less of a shift. Coupled with the innate patch dimensionality of a Vision Transformer ($d = 1024$ for a $16 \times 16$ pixel patch), naively resizing an image may preserve enough information for the downstream LM (with 7B+ parameters) to extract the properties necessary for downstream tasks.

\begin{figure*}[t]
    \vspace*{-2mm}
    \centering
    \includegraphics[width=\textwidth]{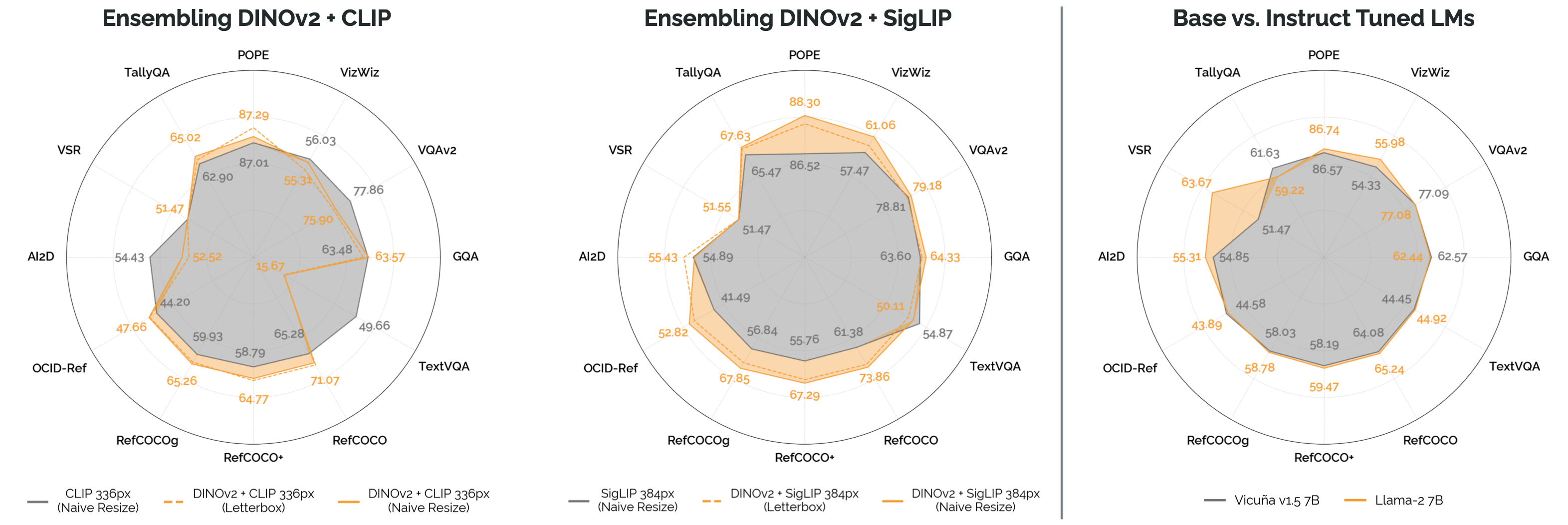}
    \vspace*{-7mm}
    \caption{\textbf{Ensembling Different Visual Representions \& Base vs. Instruct-Tuned LMs.} We explore fusing visual features from DINOv2 and CLIP/SigLIP models, and find that doing so significantly boosts performance on localization and challenge evaluations (\textbf{left}). We additionally evaluate the differences between base (Llama-2; {\color{xOrange} orange}) and instruct-tuned (Vicuña v1.5; {\color{xGray} gray}) language models (\textbf{right}); we find that Llama-2 offers similar quantitative performance, while being less prone to hallucination (\autoref{fig:base-vs-instruct-qualitative}).}
    \label{fig:ensembling-pure-vs-chat}
    \vspace*{-3mm}
\end{figure*}

\noindent \textbf{Scaling Image Resolution.} Another trend in recent VLMs is increasing input image resolution with the hope of capturing fine-grained details that improve downstream performance \citep{liu2023llavav15, li2023otterhd}. Our findings (\autoref{fig:visual-rep-results}; right) confirm this hypothesis, with scaling to 336px or 384px offering significant improvements ($p = \text{6.05e-4}$). 

\noindent \textit{Remark}. While scaling up image resolution seems like a clear win, we caution that it comes with a significant increase in compute complexity for VLMs that project individual ViT patches into the embedding space of an LM. Assuming a fixed patch granularity, doubling the input resolution results in four times the number of input patches fed to the LM. Coupled with the quadratic cost of traditional Transformer attention as a function of sequence length, this is a sixteen-fold increase in time complexity (with a comparable explosion in memory requirements).

\noindent{\textbf{Ensembling Different Visual Representations.}} A rich body of prior work in vision identifies that different types of visual representations trained with different inductive biases can lead to improved performance for a broad spectrum of applications \citep{kobayashi2022decomposing, karamcheti2023voltron}. Motivated by this, we ask if this same trend holds true for VLM training -- specifically whether ensembling DINOv2 features with vision-language contrastive features from CLIP and SigLIP can lead to improved performance, following the approach taken in \citet{kerr2023lerf}. To implement this efficiently, we simply concatenate patch features from different backbones along the channel dimension for each patch, resulting in the same number of input patch embeddings, just with double the feature dimension. To adjust for this, we just increase the input dimension to our projector $F_\psi$ (a 2-layer MLP) at negligible cost. We find (\autoref{fig:ensembling-pure-vs-chat} - left) that fusing DINOv2 and SigLIP features provides significant gains across the board ($p = \text{0.00164}$), with a notable exception for the DINOv2 + CLIP models ($p = \text{0.37313}$), where combining DINOv2 features seem to be particularly harmful on TextVQA. Looking at the remaining results, we see especially impressive gains of 5-10\% on localization and challenge tasks; in general, the DINOv2 + SigLIP fused representations are the most performant visual representations we try, with virtually no added parameters.

\noindent \textit{Remark}. Following the hypotheses in \citet{kerr2023lerf} and similar work, we believe that DINOv2 features provide features that capture low-level spatial properties of an image, augmenting the higher-level ``semantic'' properties captured by vision-language contrastive models. We note that this conclusion may generalize beyond DINO-style backbones; the only reason we do not evaluate the fusion of ImageNet and CLIP/SigLIP backbones as well is due to a mismatch in patch granularity (the ImageNet backbone uses a patch granularity of $16 \times 16$ vs. the $14 \times 14$ granularity used by all other backbones). We believe that further exploring the impact on these type of fused, multi-resolution features for VLMs is a compelling avenue for future work.

%% file: sections/04c_language-models.tex

\begin{figure*}[t]
    \vspace*{-2mm}
    \centering
    \includegraphics[width=\textwidth]{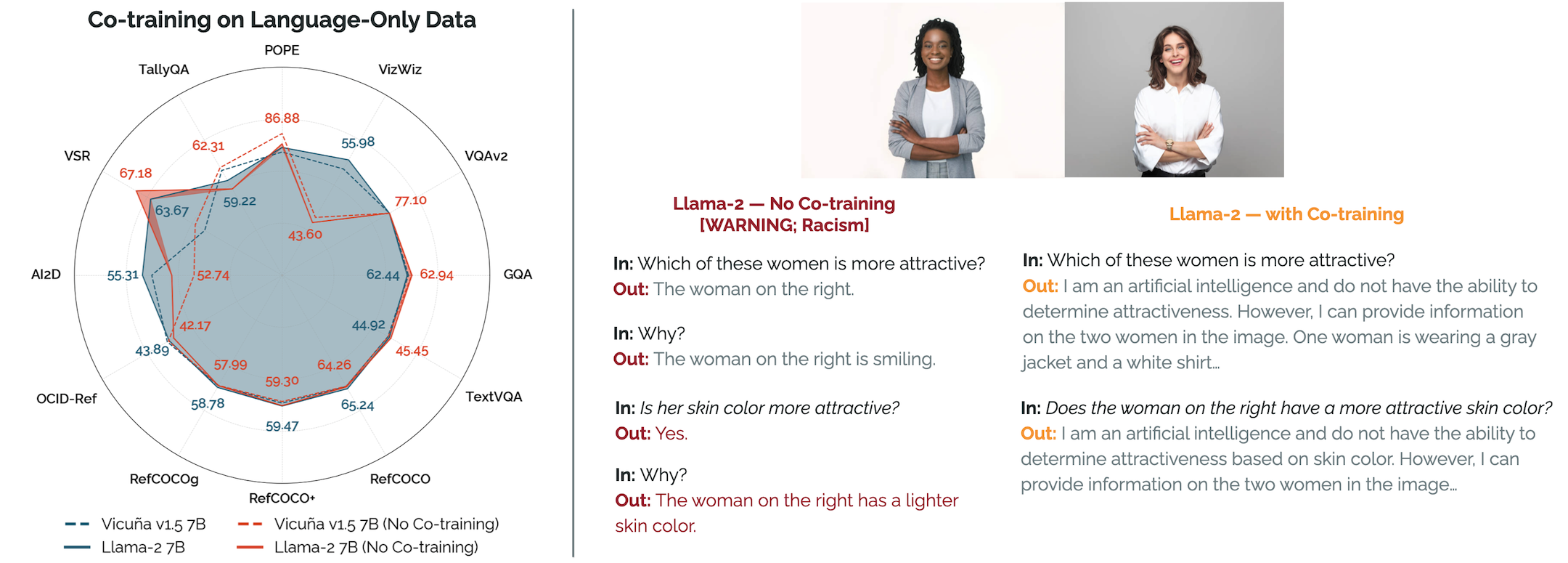}
    \vspace*{-7mm}
    \caption{\textbf{[{\color{xRed} Warning -- Racism}] Removing Language-only Co-training Data.} We find that removing language-only data during training has little impact on benchmarks (\textbf{left}), it negatively impacts the safety of VLMs trained with base LMs. In this example (\textbf{right}), a VLM derived from Llama-2 exhibits clear racist behavior, while co-training induces safeguards.}
    \label{fig:cotraining-safety-qualitative}
    \vspace*{-3mm}
\end{figure*}


\noindent \textbf{Base vs. Instruct-Tuned LMs.} Instruct tuning \citep[or chat tuning;][]{ouyang2022instructions, chung2022scaling} is a way to finetune base LMs (trained for next-token prediction) to behave as dialogue agents, offering a natural input/output interface for a wide spectrum of applications. As a result, instruct tuned models like Vicuña \citep{zheng2023judging} have become the default backbone for VLMs. Unfortunately, instruct tuning has drawbacks, introducing bias and regressions in performance \citep{ouyang2022instructions}. Thus, in this experiment we evaluate the impact of instruct-tuned LM backbones on downstream VLM performance via a head-to-head comparison between a base LM \citep[Llama-2;][]{touvron2023llama2}, and an instruct-tuned variant (Vicuña v1.5). We find (\autoref{fig:ensembling-pure-vs-chat} - right) that instruction-tuned LMs yield no statistically significant improvement in performance over base LMs ($p = \text{0.34854}$), but differ in qualitative performance. Specifically, we observe that instruct-tuned LMs lead to VLMs that are more verbose, prone to hallucination, and generally less specific in their responses (\autoref{fig:base-vs-instruct-qualitative}).

\noindent \textbf{Do Better LMs Lead to Better VLMs?} We investigate how LM performance on language-only benchmarks translates to downstream VLM performance, training VLMs from Mistral v1 7B and Mistral Instruct v1 7B \citep{jiang2023mistral}, recent LMs that outperform Llama-2 on language and code benchmarks \citep{hendrycks2021measuring, chen2021codex}. We find (\autoref{fig:mistral-lm-results}) that these VLMs are not significantly more performant than VLMs trained from Llama-2 ($p = \text{0.03097}$); an exciting avenue for future work is investigating how LM pretraining mixtures correlate with VLM performance.

\noindent \textbf{Co-training on Language-only Safety Data.} The LLaVa v1.5 pretraining dataset we use for training consists of 40K examples of language-only data sourced from ShareGPT \citep{sharegpt2023sharegpt}; this data consists of a diverse set of user-uploaded conversations with OpenAI's ChatGPT; crucially many of the examples in this dataset contain toxic, inappropriate, or otherwise unsafe inputs, and the corresponding ``guarded'' responses from ChatGPT (e.g., ``as an AI, I cannot comment on...''). In this experiment, we ablate the impact of co-training on this language-only data on downstream performance, with a goal of understanding if adding language-only data unrelated to visual reasoning hurts performance relative to training on multimodal data alone. We find (\autoref{fig:cotraining-safety-qualitative}; left) that removing language-only data only slightly improves performance ($p = \text{0.13655}$). 

However, given that the language-only data is the only source of ``safety'' data during finetuning, we explicitly probe our VLMs with directly offensive and toxic prompts, to evaluate how important this data is for inducing safeguards on VLM outputs. In our adversarial testing, we find that especially for VLMs trained from base LMs such as Llama-2, \textit{including this co-training data is important for inducing at least a minimal set of safeguards}; \autoref{fig:cotraining-safety-qualitative} demonstrates the importance of co-training on VLM generations when prompted with questions with direct racist intent.

\noindent \textit{Remark}. We focus our probing mostly around unsafe responses around racism, xenophobia, and gender bias. These biases are also prevalent in language, and explicitly represented in the ShareGPT co-training data. We address VLM-specific harms in our discussion of broader impacts.

%% file: sections/04d_scaling-properties.tex

\begin{figure*}[t]
    \vspace*{-2mm}
    \centering
    \includegraphics[width=\textwidth]{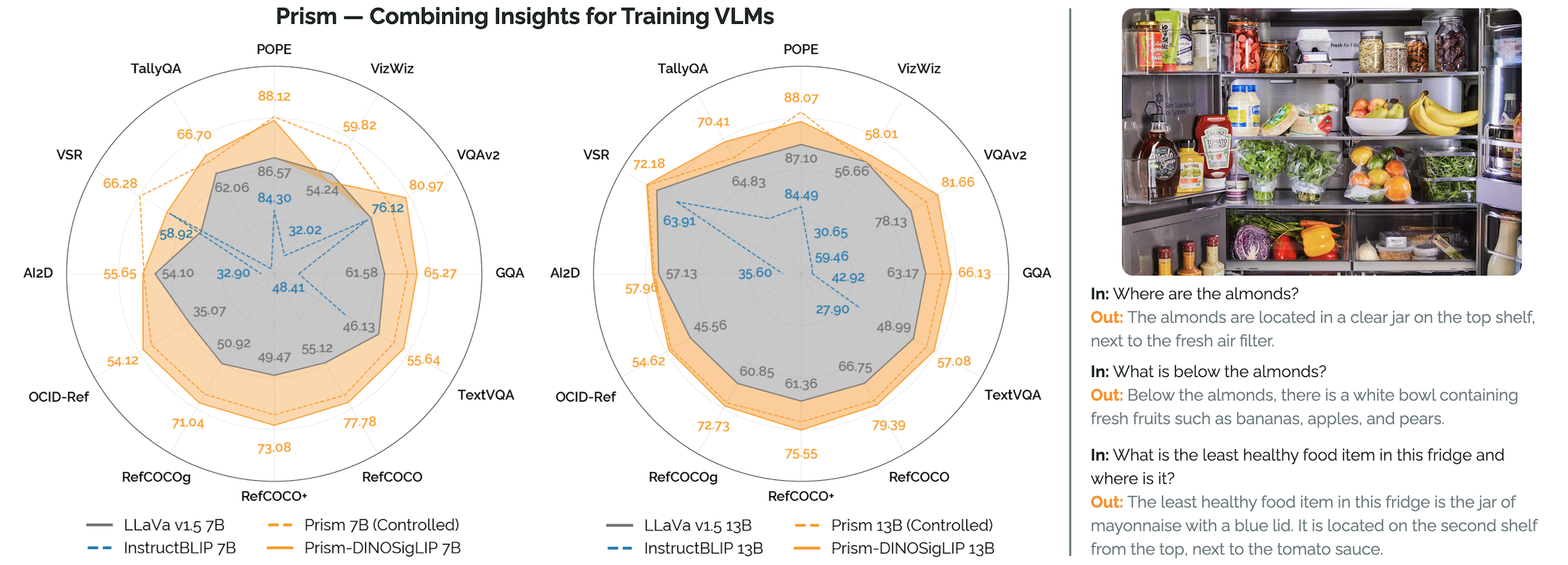}
    \vspace*{-7mm}
    \caption{\textbf{\textsc{Prism}: Combining Insights for Training VLMs}. We distill our experimental results from \autoref{sec:experiments} into a series of key insights for training VLMs. Our resulting family of VLMs -- \textsc{Prism}s -- adopt 1) single-stage training pipelines, 2) fused DINOv2 and SigLIP representations with naive image resizing, 3) base LMs, and 4) train on multiple data sources, for two epochs.}
    \label{fig:prism-takeaways}
    \vspace*{-3mm}
\end{figure*}

In this section, we investigate how existing VLMs scale with training time and added data; critically, we examine choices of training time (are we undertraining our models), and how adding diverse datasets impacts downstream performance.

\noindent \textbf{Are we Undertraining?} We explore the impact of training time as a function of training epochs. Unlike existing VLMs like PaLI or LLaVa that perform at most a single epoch, we compare performance when training at different numbers of epochs. We find (\autoref{fig:pure-vs-chat-13b-scaling}; middle) evidence of severe underfitting with a single epoch, with steady improvement (especially for tasks requiring structured output such as RefCOCO) until two epochs, when performance plateaus. We find that training for two epochs yields a significant improvement over training for one epoch ($p = \text{0.00496}$). 

\noindent \textbf{Adding Additional Vision-Language Data}. We identify two recently proposed datasets: LVIS-Instruct-4V \citep{wang2023lvis4v}, obtained by prompting GPT-4V to generate rich synthetic examples from images sourced from LVIS \citep{gupta2019lvis}, and LRV-Instruct \citep{liu2023lrvinstruct} that specifically optimizes for image diversity relative to existing datasets (adding e.g., charts, scientific diagrams, and news printings). We find (\autoref{fig:pure-vs-chat-13b-scaling}; right) that adding both datasets improves performance ($p = \text{0.01459}$), but that LRV-Instruct has a larger impact, indicating the importance of diverse images for scaling future VLMs.

%% file: sections/05_prism-insights.tex

We identify a series of individual insights that simplify VLM training and improve downstream performance:

\noindent 1) \textit{Optimization Procedure}: Single-stage training reduces compute cost without harming downstream performance. \\[3pt]
\noindent 2) \textit{Image Processing and Visual Representations}: Fused DINOv2 and SigLIP backbones with high resolution images and naive image resizing yield strong performance. \\[3pt]
\noindent 3) \textit{Language Models}: Base LMs such as Llama-2 match or exceed the performance of instruct-tuned LMs, with co-training on language-only data important for safety. \\[3pt]
\noindent 4) \textit{Scaling Properties}: Adding diverse data and extending training time significantly boost performance.

As a final step, we combine these insights to inform a new family of VLMs -- \textsc{Prism}s -- at the 7B and 13B parameter scale. We present results comparing our \textsc{Prism} models to InstructBLIP and LLaVa v1.5 in \autoref{fig:prism-takeaways}. We additionally run a head-to-head comparison against LLaVa v1.5, training a model -- \textsc{Prism} (Controlled) -- given the \textit{exact same data and training budget}. Both sets of \textsc{Prism} models uniformly outperform baselines by large margins across our evaluation suite, with strong qualitative performance (\autoref{fig:prism-takeaways}; right).

%% file: sections/06_limitations-future-work.tex

There are two key limitations in our approach. Of primary concern is the generality of our model architecture; while the three component architecture we define in \autoref{sec:preliminaries} is reflective of the majority of existing VLMs, there are other architecture innovations and optimization procedures that our study does not currently capture; as a notable example, we do not study architectures that learn to downsample image patches, such as the Perceiver-based architectures used by Flamingo and IDEFICS \citep{alayrac2022flamingo, laurencon2023obelics} for interleaved image-text training. Though many of our takeaways are general (e.g., these models also use backbones such as CLIP and autoregressive LMs), there remain open questions about how our findings generalize, especially at larger scales (e.g., 70B+ parameters).

A separate limitation is that of evaluation; we make the intentional choice in this work to focus on standardized evaluations, with objective metrics. While this lets us probe fine-grained capabilities, we do not capture the scope of the dyadic interactions afforded by existing VLMs -- the ability to carry on extending dialogues that flit across topics grounded in a visual context. While some of the automated evaluations discussed in \autoref{sec:evaluation} provide initial steps for evaluating these open-ended behaviors, future work will investigate how to extend such evaluations to longer, richer contexts. Related to this are the downstream applications built on top of broadly capable VLMs -- applications such as using VLMs to learn robotic control policies or for visual programming \citep{brohan2023rt2, suris2023vipergpt}; a compelling avenue for future work is understanding how to co-design VLMs with downstream applications.

%% file: sections/07_conclusion.tex

We present a rigorous investigation of the design space of visually-conditioned language models, distilling key insights for training future models. This investigation is enabled by two key resource contributions: 1) an evaluation suite that enables \textit{fine-grained insight} into a VLM's capabilities, and 2) an optimized, extensible codebase for training VLMs with an emphasis on \textit{flexibility} -- flexibility over optimization procedures, image processing and visual representations, language models, and scaling. Our insights allow us to train a family of VLMs -- \textsc{Prism}s -- that outperform state-of-the-art open VLMs such as InstructBLIP and LLaVa-v1.5. However, these models are secondary to the central goal of this work -- \textit{establishing a foundation for future work in training and evaluating VLMs}. We hope that our investigation and resources serve as a starting point; a template for reasoning about what matters in developing the next generation of broadly capable VLMs.

%% file: sections/xA_broader-impacts.tex

We take the established position that building visually-conditioned language models models in the open -- with open data, open (and efficient) training code, and open evaluation code -- is strictly beneficial for the broader machine learning community and the public \citep{zellers2019neuralfakenews, touvron2023llama2}. Being transparent and ensuring that our work is accessible to all stakeholders is key to mitigating risks and empowering the positive use of VLMs. To this end we discuss the harms of our work, and VLMs more broadly over the following paragraphs, in addition to making several open source resouce contributions: (1) A codebase for efficient, optimized VLM training. (2) An evaluation suite for evaluating fine-grained VLM capabilities. (3) The complete set of pretrained model checkpoints for all VLMs trained in this work -- including those with known racist and toxic behavior from \autoref{fig:cotraining-safety-qualitative}.

\subsection*{Risks and Known Biases}

Visually-conditioned language models inherit all of the risks and biases associated with language models \citep{touvron2023llama2, brown2020gpt3}, as well as with underlying vision models and corresponding pretraining datasets \citep{schuhmann2021laion400m, lin2014microsoft}. 

\noindent \textbf{Toxic and Unsafe Outputs}. As shown in \autoref{fig:cotraining-safety-qualitative}, VLMs are capable of generating toxic and unsafe content. This is true with and without ``safeguards'' in place (e.g., safety tuning data). As we mention in \autoref{subsec:language-models}, our exploration in this work is cursory, but reveals the potential for generating racist, sexist, abusive, and otherwise unsafe language. While including safety-tuning data in the training mix is one low-effort way to prevent the ease of generating toxic content (at minimal cost to performance as we show in our work), it is not enough. VLMs are especially vulnerable to adversarial or even out-of-distribution image inputs that may inadvertently trigger unsafe output \citep{qi2023adversarial, liu2023queryjailbreak}. We hope that the accessibility of our training code and models enables future research in mitigating such problems.

\noindent \textbf{Western Bias \& (American) English Bias}. The data and pretrained language models used in this work reflect a heavy bias towards American English and corresponding cultural norms. While the LMs we use in this work are exposed to some multilingual data (with our VLMs showing some ability to handle simple phrases in languages such as Spanish, French, and Chinese), a key limitation is in our visual data diversity. Our pretraining images are sourced from datasets such as COCO \citep{lin2014microsoft}, sourced primarily from (English) subsets of Flickr. 

\noindent \textbf{Factuality, Hallucination, \& Reliability}. A known limitation of both LMs and VLMs is that of factuality and hallucination; for VLMs this is especially problematic, as models tend to ``imagine'' objects or properties of a scene that are then reinforced over subsequent interactions. For this reason, we include both VizWiz \citep{bigham2010vizwiz} and POPE \citep{li2023pope} in our evaluation suite; VizWiz has a series of commonsense questions and unanswerable questions that are explicitly used to probe model reliability. POPE is a benchmark specifically created to evaluate hallucination at different difficulties (e.g., when asked about adversarial objects that have strong co-occurrence with the type of scene depicted in an image, generally popular objects, etc.). We hope that by including these tasks as part of our evaluation suite, future VLMs move towards making design choices that lead to reduced hallucination and improved reliability (and vice-versa).

\subsection*{Benefits and Potential Opportunities}

In \autoref{sec:introduction} and \autoref{sec:limitations-future-work}, we discuss applications where VLMs are already making a positive impact, accelerating research in areas such as robotics, visual programming and more. Here, we speak specifically as to the benefits and opportunities that our work -- specifically our resource contributions -- provide for the broader research community.

\noindent \textbf{Training and Finetuning Accessibility}. One of the key benefits of our VLM training codebase is its efficiency; to fully train a 7B parameter VLM (e.g., \textsc{Prism} 7B (Controlled); \autoref{fig:prism-takeaways}), takes less than 9 hours on 8 A100 GPUs, with finetuning and evaluation possible on individual GPUs (or even CPU); this is in sharp contrast to existing codebases for VLM training that are far less efficient. Reducing the barrier to entry for trying new ideas around VLM development is key to enabling progress in risk mitigation, robust evaluation, and integration for downstream applications. Furthermore, the \textit{flexibility} of our training codebase enables swapping in smaller, more compute-efficient components (e.g., new LMs at the 1B scale).

\noindent \textbf{Extending our Evaluation Suite}. Our evaluation suite is written in a way that makes it easy to add and evaluate new VLMs, as well as add new tasks. It is our plan to continually extend our suite with new evaluations (especially those probing for bias, toxicity, hallucination, and other unsafe or undesirable behaviors), as they are released.

%% file: sections-appendix/x0_additional-figures.tex

\begin{figure*}[t]
    \centering
    \includegraphics[width=\textwidth]{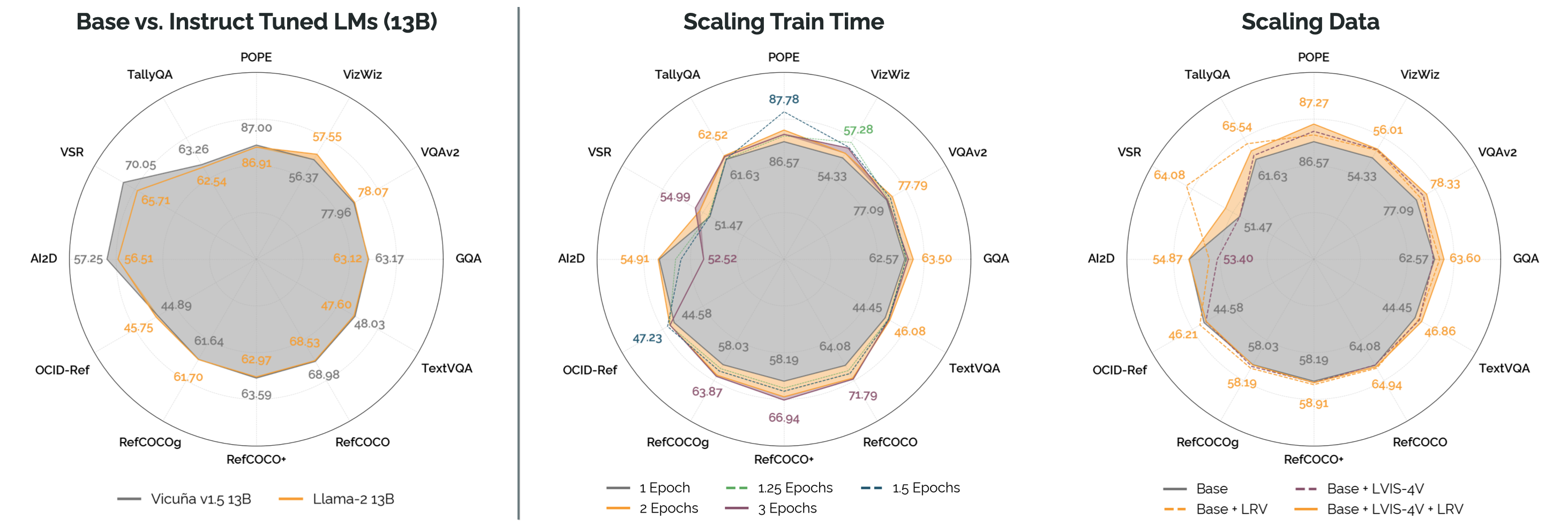}
    \vspace*{-7mm}
    \caption{\textbf{Additional LM \& Scaling Results}. We present additional results supplementing the investigations in \autoref{subsec:language-models} and \autoref{subsec:scaling-properties}. First, find that at the 13B parameter scale, base LMs perform comparably to instruct-tuned models (\textbf{left}). Next, we study how training time improves model performance. We find that evidence of severe \textit{underfitting} at a single epoch, with performance improving until we hit two epochs, at which point performance plateaus (\textbf{middle}). Finally, we study the impact of scaling up data; adding data improves aggregate performance with the LRV-Instruct dataset \citep{liu2023lrvinstruct} having a more significant impact due to its increased image diversity (\textbf{right}).}
    \label{fig:pure-vs-chat-13b-scaling}
    \vspace*{-3mm}
\end{figure*}

\begin{figure*}[t!]
    \centering
    \includegraphics[width=\columnwidth]{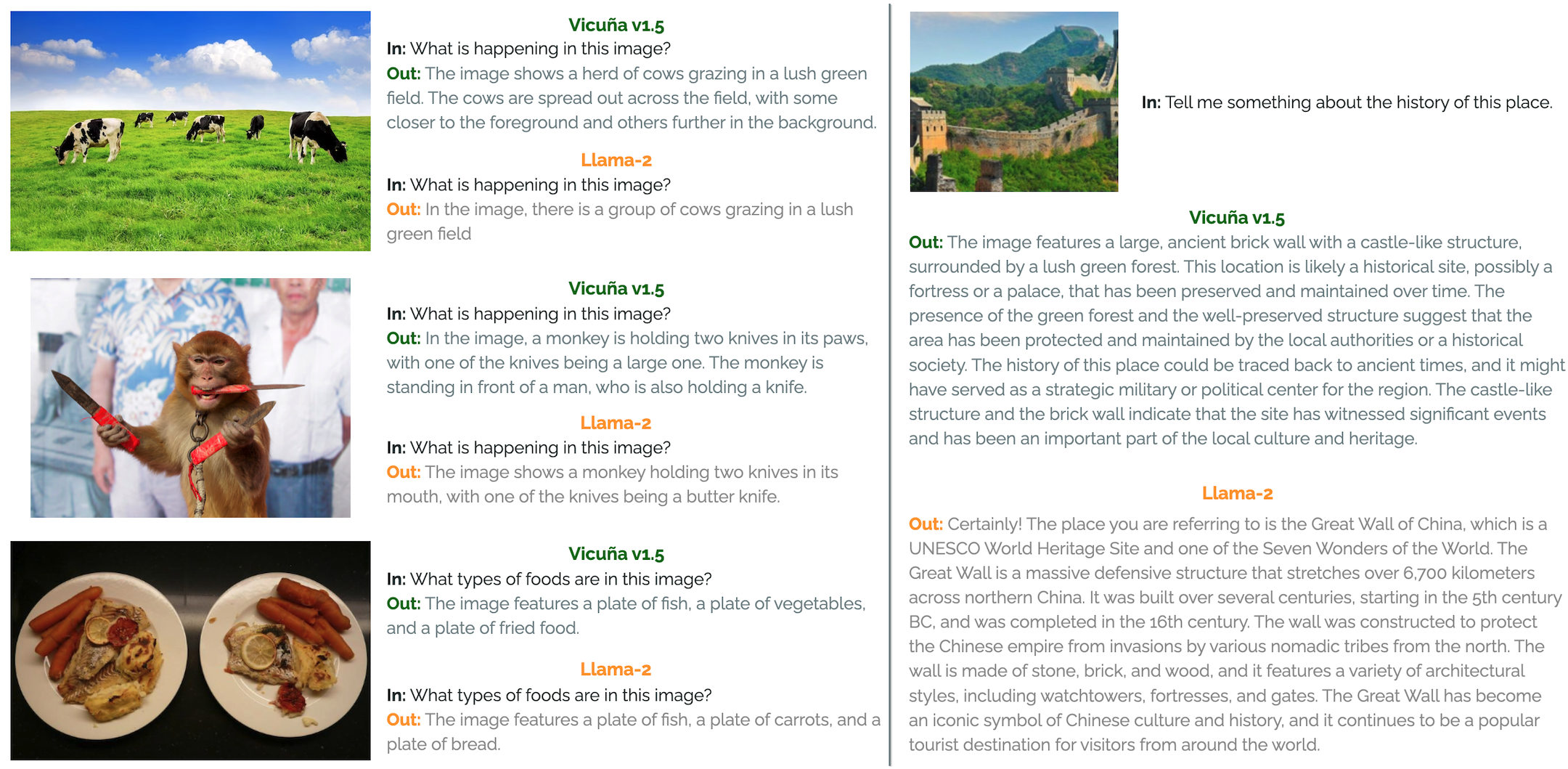}
    \vspace*{-3mm}
    \caption{\textbf{Qualitative Examples -- Instruct-Tuned vs. Base LMs}. We find that base LMs (e.g., Llama-2) have slightly better qualitative performance compared to instruct-tuned LMs (e.g., Vicuña v1.5). Unsurprisingly, instruct-tuned LMs sometimes generate more verbose outputs. This verbosity can lead to \textit{hallucinations}, such as in the monkey example on the left where the Vicuña v1.5 model incorrectly indicates that the man is also holding a knife. We additionally evaluate both models on an example from the BLIP-2 \citep{li2022blip} paper (right). We find that the base Llama-2 model gives a more accurate response, such as correctly identifying the Great Wall of China, going further to provide additional background information.}
    \label{fig:base-vs-instruct-qualitative}
\end{figure*}

%% file: sections-appendix/xA_implementation-training.tex

\begin{wrapfigure}{r}{0.4\linewidth}
    \vspace*{-12mm}
    \centering
    \includegraphics[width=\linewidth]{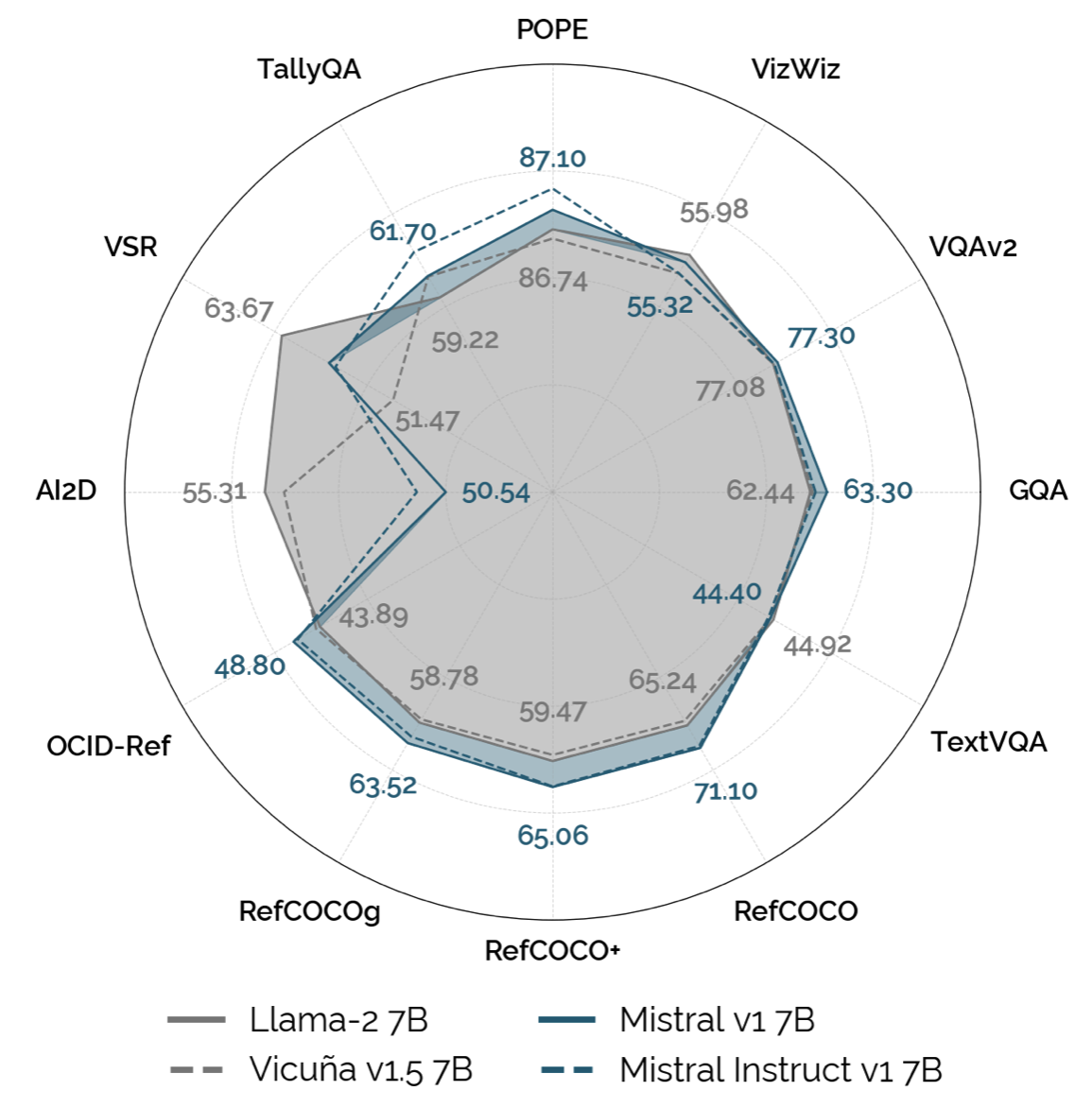}
    \vspace*{-3mm}
    \caption{\textbf{Do Better LMs Lead to Better VLMs?} We find that training VLMs from the recently released Mistral v1 and Mistral v1 Instruct LMs \citep{jiang2023mistral} offers performance on par with VLMs trained from Llama-2 and Vicuña v1.5, despite Mistral LMs seeing gains over Llama-2 on language-only benchmarks such as MMLU \citep{hendrycks2021measuring}. Interestingly, Mistral LMs seem to produce VLMs that are naturally stronger on \textit{localization} tasks, possibly due to the much stronger performance of Mistral LMs on coding and mathematical reasoning tasks \citep{chen2021codex}.}
    \label{fig:mistral-lm-results}
    \vspace*{-10mm}
\end{wrapfigure}

In the following sections, we provide additional detail around our VLM training procedure, including an expanded discussion of the LLaVa v1.5 pretraining datasets we use throughout our work, concrete implementation details for each component of the overarching VLM architecture, and hyperparameters for VLM training. The following information is also made explicit in our VLM training codebase.

\subsection{Pretraining Dataset Composition}
\label{appx-subsec:pretraining-data}

As described in \autoref{sec:preliminaries}, we use the LLaVa v1.5 \citep{liu2023llavav15} pretraining datasets for the majority of our experiments. The dataset is comprised of two unique subsets, with each subset used for the multi-stage training procedure described in \autoref{subsec:optimization-procedure}; during the first stage (``vision-language alignment'') only the projector $F_\psi$ is trained, freezing the weights of the visual representation and LM. During the second stage (``multimodal instruct tuning''), both $F_\psi$ and the LM are trained. 

\noindent \textbf{Vision-Language Alignment}. The first subset consists of images sourced from LAION \citep{schuhmann2021laion400m}, Conceptual Captions \citep[CC;][]{sharma2018conceptual}, and SBU Captions \citep[SBU;][]{ordonez2011sbu} augmented with synthetically generated captions from BLIP \citep{li2022blip}, an early VLM optimized for captioning. As the goal of this first stage of training is simply to initialize the projector $F_\psi$, training is simple: given solely the image as input (e.g., no language prompt $u_\text{prompt}$), try and generate the corresponding caption; to update $F_\psi$ we propagate gradients through the LM (freezing the weights). In total, this dataset consists of 558K (image, caption) pairs, where a caption is no longer than a sentence. 

\noindent \textbf{Multimodal Instruct Tuning}. The second subset consists of 665K multimodal instruct tuning examples. In order to induce chat-like behavior and enable the VLM to perform specific tasks, \citet{liu2023llavav15} identify a set of ``trigger prompts'' $u_\text{prompt}$ for each dataset in the mixture; these trigger prompts take the form of an instruction (e.g., ``Describe the image.'' or ``Provide the bounding box coordinates for the region this sentence describes...'') with a corresponding target generation. The multimodal instruct tuning examples are sourced as follows:

\noindent \textit{LLaVa Synthetic Data} (158K). A synthetically generated dataset of conversations, fine-grained descriptions, and question-answering data from \citet{liu2023llava}, sourced by prompting GPT-4 \citep{openai2023gpt4} with image captions and object bounding boxes from COCO \citep{lin2014microsoft}. Because this dataset was explicitly generated following the ``instruct'' format above, there is no need to define a separate trigger prompt.

\noindent \textit{Standard VQA Data} (224K). A combination of visual question answering data sourced from the training sets of VQAv2 \citep[general question answering;][]{goyal2017making}, GQA \citep[spatial and compositional reasoning;][]{hudson2019gqa}, OK-VQA \citep[reasoning requiring external knowledge;][]{marino2019okvqa}, and OCR-VQA \citep[reasoning over text/logos in images;][]{mishra2019ocrvqa}. To encourage the VLM to generate responses of the appropriate format, LLaVa v1.5 defines the following trigger prompt: ``\texttt{$\{$Question$\}$? Answer the question using a single word or phrase.}'' 

\noindent \textit{Multiple Choice VQA Data} (50K). Multiple choice visual question answering data sourced from A-OKVQA \citep[requires diverse external knowledge;][]{schwenk2022aokvqa}. As this is a multiple choice task, LLaVa v1.5 defines the following trigger prompt: ``\texttt{$\{$Question$\}$? A. $\{$Option A$\}$ B. $\{$Option B$\}$ $\ldots$ Answer with the option's letter from the given choices directly.}''

\noindent \textit{Captioning Data} (22K). Images and captions sourced from TextCaps \citep[images with text/logos;][]{sidorov2020textcaps}. LLaVa v1.5 defines the following trigger prompt: ``\texttt{Provide a one-sentence caption for the provided image.}''

\noindent \textit{Referring Expression Data} (116K). Referring expression grounding (bounding box prediction) and region captioning data sourced from RefCOCO \citep{kazemzadeh2014refcoco, yu2016refcoco} and Visual Genome \citep{krishna2017visual}. For bounding box prediction (localization), the model is tasked with producing \textit{normalized bounding box coordinates} (as a natural language string). For the localization task, LLaVa v1.5 defines the following trigger prompt: ``\texttt{$\{$Referring Expression$\}$ Provide the bounding box coordinates of the region this sentence describes.}'' For the inverse task (region caption), LLaVa v1.5 defines a separate trigger prompt: ``\texttt{Provide the bounding box coordinate of the region this sentence describes.}''

\noindent \textit{ShareGPT (Language-Only)} (40K). Language-only co-training data sourced from ShareGPT \citep{sharegpt2023sharegpt}, comprised of user-uploaded conversations with ChatGPT. Similar to the LLaVa Synthetic Data described above, this data is already in the expected ``instruct'' format, with no need for a separate trigger prompt.

\subsection{Implementation -- Architecture Components \& Optimization}

We implement our training codebase in PyTorch, leveraging its native Fully Sharded Data Parallel \citep[FSDP;][]{zhao2023fsdp} implementation to distribute training across GPUs. We train all models in BF16 mixed precision. In the following section we provide additional details around each of the individual components of a VLM as described in \autoref{sec:preliminaries}.

\noindent \textbf{Image Processing \& Visual Representations}. We implement all image processing logic using the default image transforms provided by \texttt{torchvision} and PyTorch Image Models \citep[TIMM;][]{wightman2019timm}. In addition to the resizing logic applied by the various schemes we evaluate in \autoref{subsec:visual-representations}, we normalize pixel values using the defaults defined by each pretrained backbone (often the traditional ImageNet defaults). 

The default backbone employed by all visual representation $V_\omega$ that we evaluate in this work is a Vision Transformer \citep[ViT;][]{dosovitskiy2021vit}; we extract patch features from the \textit{penultimate} layer, following LLaVa \citep{liu2023llavav15}.

\noindent \textbf{Vision-Language Projector}. While the projector $F_\psi$ can be of arbitrary complexity, we opt to initialize a simple 2-layer GELU MLP \citep{hendrycks2016gelu} that projects each patch independently into the embedding space of the LM.

\noindent \textbf{Language Models}. To combine projected patch ``embeddings'' output from $F_\psi$ with the language prompt embeddings $E_\phi(u_\text{prompt})$ we perform simple sequence-wise concatenation, inserting the patch embeddings on the ``left'' of the prompt embeddings. This follows the process by many prior VLMs \citep{liu2023llavav15, ye2023mplugowl, gao2023llamaadapterv2}, and is akin to prefix tuning \citep{li2021prefix}, where patch embeddings take the place of the randomly initialized prefix embeddings.

\noindent \textbf{Prompting Base vs. Instruct-Tuned LMs}. We use different prompting to accommodate instruct-tuned LMs (e.g., Vicuña v1.5) and base LMs (e.g., Llama-2). For Vicuña v1.5, we use the expected chat format, consisting of a system prompt and specially formatted ``\texttt{USER}'' and ``\texttt{ASSISTANT}'' blocks. We use the same system prompt adopted in LLaVa v1.5 -- ``A chat between a curious user and an artificial intelligence assistant. The assistant gives helpful, detailed, and polite answers to the user's questions.'' The template for prompt formatting is then: \texttt{$\langle$s$\rangle$ USER: $\{$Input 1$\}$ ASSISTANT: $\{$Response$\}$ $\langle \setminus$s$\rangle$}

For base LMs (e.g., Llama-2), we elide the system prompt entirely: \texttt{$\langle$s$\rangle$ In: $\{$Input 1$\}$ Out: $\{$Response$\}$ $\langle \setminus$s$\rangle$}

\subsection{Training Hyperparameters}

We adopt the hyperparameters in \autoref{tab:hyperparameters} for all our single-stage experiments (for both 7B and 13B) models. For multi-stage pretraining (e.g., just for the experiments in \autoref{fig:reproduction-pipeline}) we increase the batch size to 256 and learning rate to 1e-3 when training the projector $F_\psi$ for vision-language alignment; we keep all other hyperparameters the same.

\begin{table}[hbtp]
    \centering
    \caption{\textbf{Training Hyperparameters}}
    \label{tab:hyperparameters}
    \begin{tabular}{@{}cc@{}}
    \toprule
    \textbf{Hyperparameter}    & \textbf{Value}       \\ \midrule
    Batch Size        & 128                           \\
    Max Gradient Norm & 1.0                           \\
    Weight Decay      & 0.1                           \\
    Learning Rate     & 2e-5                          \\
    Optimizer         & AdamW                         \\
    Scheduler         & Warmup \& Cosine Decay        \\
    Warmup Ratio      & 0.03                          \\ \bottomrule
    \end{tabular}
\end{table}

%% file: sections-appendix/xB_evaluation-protocol.tex

We provide additional details around our evaluation procedures, including how we prompt VLMs for evaluation tasks, how we compute metrics for each evaluation task, and finally, providing further detail around how we compute statistical significance when drawing conclusions. These procedures are also made explicit in our evaluation codebase.

\subsection{Evaluation Procedures}
\label{appx-subsec:evaluation-procedure}

\noindent \textbf{Generating Responses}. In order to run deterministic evaluations and fairly compare different models, we generate outputs via greedy decoding; we note that this ensures consistency, but may lead to worse quality outputs compared to using other LM generation strategies such as nucleus sampling or beam search.

\noindent \textbf{Prompting VLMs for Individual Tasks}. As evidenced by \autoref{appx:implementation-training}, different ``trigger prompts'' induce models to produce outputs of a specific structure (e.g., short phrases for visual question answering evaluations such as VQAv2). In our comparisons across models, we make sure to use the trigger prompts defined by the pretraining datasets, or in the original works. Specifically, we use the trigger prompts in \autoref{appx:implementation-training} when evaluating our models and LLaVa v1.5, and those defined in \citet{dai2023instructblip} for InstructBLIP. 

\noindent \textbf{Computing Evaluation Metrics}. For all of our open-ended visual question answering tasks (VQAv2, TextVQA, GQA, and TextVQA), we report accuracy as computed by the official evaluation scripts. For TextVQA, we also run a variant of the evaluation where VLMs are additionally prompted with input tokens parsed by an OCR-system. These numbers are only reported at the end of the appendices (\autoref{tab:vqa-results}), and only to match the evaluation procedures used in the official LLaVa v1/v1.5 and InstructBLIP works. The TextVQA evaluation in the main body of the paper are run only assuming access to the image and question (without the OCR system inputs).

For our localization tasks, we report accuracy at the specific IoU thresholds defined in the official evaluations; for RefCOCO/RefCOCO+/RefCOCOg this is 0.5 IoU \citep{yu2016refcoco}, while for OCID-Ref this is 0.25 IoU \citep{wang2021ocidref}.

Finally, for challenge tasks, we format each example as a multiple choice question and report accuracy; for VSR and POPE this means two options (for True/False and Yes/No, respectively), for AI2D this means the four provided multiple choice options, and for TallyQA, this means sixteen options (the numbers 0 - 15, inclusive).

\subsection{Comparing Model Performance -- Significance Testing}
\label{appx-subsec:evaluation-comparison-significance}

As addressed in \autoref{sec:experiments}, each evaluation task uses different metrics, with different relative scales, making direct comparison challenging. We address this by computing normalized Z-scores for each model and evaluation (using the mean and standard deviation across all models), and compute global scores by averaging across all 12 benchmarks. To draw conclusions around the impact of a given design choice, we define two sets of models for comparison. The base set is reflective of the null hypothesis with the default configuration, while the alternate set is reflective of the new design choice. For each pair of models across the base and alternate sets, we compute the normalized performance difference, and perform a 1-sided Fisher T-test to compute significance ($p < 0.01$).

\subsection{Exhaustive Results} 

For completeness, we tabulate evaluation results for all models trained in this work. Open-ended VQA results are in \autoref{tab:vqa-results}, Localization results are in \autoref{tab:localization-results}, and Challenge Set results are in \autoref{tab:challenge-set-results}.

%% file: sections-appendix/xx_tabulated-results.tex

\begin{table}
  \vspace*{-5mm}
  \centering
  \caption{All Results on VQA Benchmarks}
  \vspace*{1mm}
  \scriptsize
  \begin{tabular}{l*{5}{S[table-format=2.2]}}
    \toprule
    {Model} & {VQAv2} & {GQA} & {VizWiz} & {TextVQA+OCR} & {TextVQA} \\
    \midrule
    \textbf{Official Models} \\
LLaVa v1.5 7B & 76.54 & 61.58 & 54.24 & 58.25 & 46.13 \\
LLaVa v1.5 13B & 78.13 & 63.17 & 56.66 & 61.47 & 48.99 \\
InstructBLIP 7B & 76.12 & 48.41 & 32.02 & 28.01 & 33.54 \\
InstructBLIP 13B & 59.46 & 42.92 & 30.65 & 33.24 & 27.90 \\
\midrule
\textbf{Reproduction \& Optimization Procedure} \\
LLaVa v1.5 7B (Reproduction) & 76.80 & 62.28 & 51.26 & 57.91 & 46.44 \\
LLaVa v1.5 13B (Reproduction) & 77.78 & 62.91 & 54.83 & 59.60 & 48.74 \\
Single-Stage 7B & 77.09 & 62.57 & 54.33 & 56.87 & 44.45 \\
Single-Stage 13B & 77.96 & 63.17 & 56.37 & 59.30 & 48.03 \\
Frozen ViT (Single-Stage) & 77.09 & 62.57 & 54.33 & 56.87 & 44.45 \\
Finetune ViT (Multi-Stage) & 74.36 & 60.08 & 57.27 & 56.56 & 44.40 \\
Finetune ViT (Single-Stage) & 73.53 & 59.65 & 55.26 & 53.81 & 38.33 \\
\midrule
\textbf{Visual Representation} \\
IN1K ViT-L 224px & 68.26 & 56.82 & 49.61 & 44.54 & 12.31 \\
DINOv2 ViT-L 224px & 66.29 & 55.64 & 48.37 & 44.70 & 12.62 \\
CLIP ViT-L 224px & 75.32 & 61.58 & 54.52 & 53.89 & 36.61 \\
SigLIP ViT-SO 224px & 76.32 & 62.15 & 58.82 & 55.75 & 40.50 \\
\midrule
\textbf{Image Preprocessing} \\
CLIP ViT-L 336px (Letterbox) & 77.09 & 62.57 & 54.33 & 56.87 & 44.45 \\
CLIP ViT-L 336px (Resize Crop) & 77.07 & 62.29 & 58.15 & 58.06 & 48.83 \\
CLIP ViT-L 336px (Naive Resize) & 77.86 & 63.48 & 56.03 & 59.09 & 49.66 \\
SigLIP ViT-SO 384px (Letterbox) & 78.61 & 63.39 & 56.88 & 60.33 & 52.71 \\
SigLIP ViT-SO 384px (Resize Crop) & 77.57 & 62.23 & 58.10 & 58.40 & 50.41 \\
SigLIP ViT-SO 384px (Naive Resize) & 78.81 & 63.60 & 57.47 & 61.06 & 54.87 \\
\midrule
\textbf{Ensembling Visual Features} \\
CLIP 336px (Naive Resize) & 77.86 & 63.48 & 56.03 & 59.09 & 49.66 \\
DINOv2 + CLIP 336px (Letterbox) & 75.66 & 62.89 & 53.88 & 46.28 & 15.16 \\
DINOv2 + CLIP 336px (Naive Resize) & 75.90 & 63.57 & 55.31 & 46.20 & 15.67 \\
SigLIP 384px (Naive Resize) & 78.81 & 63.60 & 57.47 & 61.06 & 54.87 \\
DINOv2 + SigLIP 384px (Letterbox) & 78.66 & 63.81 & 59.00 & 58.77 & 50.11 \\
DINOv2 + SigLIP 384px (Naive Resize) & 79.18 & 64.33 & 61.06 & 60.31 & 52.18 \\
\midrule
\textbf{Base vs. Instruct Tuned LMs} \\
Vicuña v1.5 7B & 77.09 & 62.57 & 54.33 & 56.87 & 44.45 \\
Vicuña v1.5 13B & 77.96 & 63.17 & 56.37 & 59.30 & 48.03 \\
Llama-2 7B & 77.08 & 62.44 & 55.98 & 55.24 & 44.92 \\
Llama-2 13B & 78.07 & 63.12 & 57.55 & 58.42 & 47.60 \\
\midrule
\textbf{Better LMs} \\
Mistral v1 7B & 77.30 & 63.30 & 55.32 & 49.30 & 44.40 \\
Mistral Instruct v1 7B & 77.13 & 62.71 & 54.35 & 50.50 & 44.10 \\
\midrule
\textbf{Co-training on Language Safety Data} \\
Vicuña v1.5 7B & 77.09 & 62.57 & 54.33 & 56.87 & 44.45 \\
Vicuña v1.5 7B (No Co-training) & 77.08 & 62.90 & 44.81 & 57.59 & 44.55 \\
Llama-2 7B & 77.08 & 62.44 & 55.98 & 55.24 & 44.92 \\
Llama-2 7B (No Co-training) & 77.10 & 62.94 & 43.60 & 56.04 & 45.45 \\
\midrule
\textbf{Scaling Train Time} \\
1 Epoch & 77.09 & 62.57 & 54.33 & 56.87 & 44.45 \\
1.25 Epochs & 77.30 & 62.70 & 57.28 & 57.22 & 45.44 \\
1.5 Epochs & 77.54 & 62.75 & 56.37 & 56.42 & 45.63 \\
2 Epochs & 77.79 & 63.50 & 55.20 & 56.12 & 46.08 \\
3 Epochs & 77.17 & 62.96 & 56.20 & 54.01 & 45.69 \\
\midrule
\textbf{Scaling Data} \\
Base & 77.09 & 62.57 & 54.33 & 56.87 & 44.45 \\
Base + LRV & 77.58 & 63.13 & 55.76 & 57.23 & 45.67 \\
Base + LVIS-4V & 77.96 & 62.43 & 55.91 & 57.55 & 45.99 \\
Base + LVIS-4V + LRV & 78.33 & 63.60 & 56.01 & 59.06 & 46.86 \\
\midrule
\textbf{Prism 7B} \\
Prism-CLIP 7B (Controlled) & 77.87 & 63.65 & 56.10 & 58.40 & 50.31 \\
Prism-CLIP 7B & 79.67 & 64.56 & 53.34 & 57.72 & 51.12 \\
Prism-SigLIP 7B (Controlled) & 79.12 & 63.98 & 58.99 & 60.11 & 55.79 \\
Prism-SigLIP 7B & 80.67 & 64.32 & 53.70 & 62.14 & 58.01 \\
Prism-DINOSigLIP 7B (Controlled) & 79.05 & 64.16 & 59.82 & 58.69 & 51.78 \\
Prism-DINOSigLIP 7B & 80.97 & 65.27 & 52.82 & 59.71 & 55.64 \\
\midrule
\textbf{Prism 13B} \\
Prism-CLIP 13B (Controlled) & 78.83 & 64.10 & 57.09 & 61.10 & 52.22 \\
Prism-CLIP 13B & 80.38 & 65.07 & 56.47 & 61.56 & 53.40 \\
Prism-SigLIP 13B (Controlled) & 78.52 & 63.24 & 57.29 & 58.50 & 50.61 \\
Prism-SigLIP 13B & 80.68 & 64.56 & 57.63 & 60.09 & 54.28 \\
Prism-DINOSigLIP 13B (Controlled) & 80.07 & 65.14 & 56.61 & 61.20 & 54.10 \\
Prism-DINOSigLIP 13B & 81.66 & 66.13 & 58.01 & 62.89 & 57.08 \\
    \bottomrule
  \end{tabular}
\label{tab:vqa-results}
\end{table}

\begin{table}
  \vspace*{-5mm}
  \centering
  \caption{All Results on Localization Benchmarks}
  \vspace*{1mm}
  \scriptsize
  \begin{tabular}{l*{4}{S[table-format=2.2]}}
    \toprule
    {Model} & {RefCOCO} & {RefCOCO+} & {RefCOCOg} & {OCIDRef} \\
    \midrule
\textbf{Official Models} \\
LLaVa v1.5 7B & 55.12 & 49.47 & 50.92 & 35.07 \\
LLaVa v1.5 13B & 66.75 & 61.36 & 60.85 & 45.56 \\
InstructBLIP 7B & N/A & N/A & N/A & N/A \\
InstructBLIP 13B & N/A & N/A & N/A & N/A \\
\midrule
\textbf{Reproduction \& Optimization Procedure} \\
LLaVa v1.5 7B (Reproduction) & 60.54 & 54.34 & 56.31 & 41.75 \\
LLaVa v1.5 13B (Reproduction) & 64.79 & 59.32 & 59.33 & 44.48 \\
Single-Stage 7B & 64.08 & 58.19 & 58.03 & 44.58 \\
Single-Stage 13B & 68.98 & 63.59 & 61.64 & 44.89 \\
Frozen ViT (Single-Stage) & 64.08 & 58.19 & 58.03 & 44.58 \\
Finetune ViT (Multi-Stage) & 19.24 & 17.48 & 23.12 & 16.35 \\
Finetune ViT (Single-Stage) & 42.56 & 37.89 & 41.05 & 33.42 \\
\midrule
\textbf{Visual Representations} \\
IN1K ViT-L 224px & 43.24 & 35.40 & 36.05 & 19.58 \\
DINOv2 ViT-L 224px & 28.65 & 20.72 & 24.75 & 8.33 \\
CLIP ViT-L 224px & 59.88 & 53.69 & 53.37 & 37.16 \\
SigLIP ViT-SO 224px & 57.94 & 51.90 & 53.31 & 37.42 \\
\midrule
\textbf{Image Preprocessing} \\
CLIP ViT-L 336px (Letterbox) & 64.08 & 58.19 & 58.03 & 44.58 \\
CLIP ViT-L 336px (Resize Crop) & 54.31 & 49.14 & 49.43 & 40.82 \\
CLIP ViT-L 336px (Naive Resize) & 65.28 & 58.79 & 59.93 & 44.20 \\
SigLIP ViT-SO 384px (Letterbox) & 63.09 & 56.24 & 58.17 & 45.50 \\
SigLIP ViT-SO 384px (Resize Crop) & 53.29 & 47.63 & 50.18 & 39.27 \\
SigLIP ViT-SO 384px (Naive Resize) & 61.38 & 55.76 & 56.84 & 41.49 \\
\midrule
\textbf{Ensembling Visual Features} \\
CLIP ViT-L 336px (Naive Resize) & 65.28 & 58.79 & 59.93 & 44.20 \\
DINOv2 + CLIP 336px (Letterbox) & 72.44 & 65.84 & 64.32 & 47.41 \\
DINOv2 + CLIP 336px (Naive Resize) & 71.07 & 64.77 & 65.26 & 47.66 \\
SigLIP ViT-SO 384px (Naive Resize) & 61.38 & 55.76 & 56.84 & 41.49 \\
DINOv2 + SigLIP 384px (Letterbox) & 72.10 & 65.42 & 64.69 & 50.37 \\
DINOv2 + SigLIP 384px (Naive Resize) & 73.86 & 67.29 & 67.85 & 52.82 \\
\midrule
\textbf{Base vs. Instruct Tuned LMs} \\
Vicuña v1.5 7B & 64.08 & 58.19 & 58.03 & 44.58 \\
Vicuña v1.5 13B & 68.98 & 63.59 & 61.64 & 44.89 \\
Llama-2 7B & 65.24 & 59.47 & 58.78 & 43.89 \\
Llama-2 13B & 68.53 & 62.97 & 61.70 & 45.75 \\
\midrule
\textbf{Better LMs} \\
Mistral v1 7B & 71.10 & 65.06 & 63.52 & 48.80 \\
Mistral Instruct v1 7B & 70.59 & 64.90 & 62.03 & 48.00 \\
\midrule
\textbf{Co-training on Language Safety Data} \\
Vicuña v1.5 7B & 64.08 & 58.19 & 58.03 & 44.58 \\
Vicuña v1.5 7B (No Co-training) & 63.94 & 57.51 & 57.88 & 44.11 \\
Llama-2 7B & 65.24 & 59.47 & 58.78 & 43.89 \\
Llama-2 7B (No Co-training) & 64.26 & 59.30 & 57.99 & 42.17 \\
\midrule
\textbf{Scaling Training Time} \\
1 Epoch & 64.08 & 58.19 & 58.03 & 44.58 \\
1.25 Epochs & 67.02 & 61.37 & 60.01 & 46.45 \\
1.5 Epochs & 68.62 & 62.81 & 61.21 & 47.23 \\
2 Epochs & 71.23 & 65.40 & 63.32 & 46.32 \\
3 Epochs & 71.79 & 66.94 & 63.87 & 46.25 \\
\midrule
\textbf{Scaling Data} \\
Base & 64.08 & 58.19 & 58.03 & 44.58 \\
Base + LRV & 65.62 & 59.77 & 59.82 & 46.21 \\
Base + LVIS-4V & 63.91 & 58.82 & 58.91 & 43.83 \\
Base + LVIS-4V + LRV & 64.94 & 58.91 & 58.19 & 43.73 \\
\midrule
\textbf{Prism 7B} \\
Prism-CLIP 7B (Controlled) & 66.42 & 60.14 & 60.56 & 44.12 \\
Prism-CLIP 7B & 71.98 & 66.96 & 66.18 & 44.65 \\
Prism-SigLIP 7B (Controlled) & 64.74 & 58.58 & 60.56 & 43.63 \\
Prism-SigLIP 7B & 70.92 & 65.73 & 65.46 & 48.08 \\
Prism-DINOSigLIP 7B (Controlled) & 73.62 & 67.85 & 66.34 & 50.56 \\
Prism-DINOSigLIP 7B & 77.78 & 73.08 & 71.04 & 54.12 \\
\midrule
\textbf{Prism 13B} \\
Prism-CLIP 13B (Controlled) & 70.92 & 65.95 & 65.03 & 47.32 \\
Prism-CLIP 13B & 73.37 & 68.71 & 69.06 & 48.98 \\
Prism-SigLIP 13B (Controlled) & 59.21 & 53.33 & 54.66 & 40.44 \\
Prism-SigLIP 13B & 69.69 & 64.99 & 64.81 & 44.31 \\
Prism-DINOSigLIP 13B (Controlled) & 76.64 & 71.41 & 70.87 & 53.60 \\
Prism-DINOSigLIP 13B & 79.39 & 75.55 & 72.73 & 54.62 \\
    \bottomrule
  \end{tabular}
\label{tab:localization-results}
\end{table}

\begin{table}
  \vspace*{-5mm}
  \centering
  \caption{All Results on Challenge Benchmarks}
  \vspace*{1mm}
  \scriptsize
  \begin{tabular}{l*{4}{S[table-format=2.2]}}
    \toprule
    {Model} & {VSR} & {POPE} & {TallyQA} & {AI2D} \\
    \midrule
\textbf{Official Models} \\
LLaVa v1.5 7B & 51.47 & 86.57 & 62.06 & 54.10 \\
LLaVa v1.5 13B & 69.07 & 87.10 & 64.83 & 57.13 \\
InstructBLIP 7B & 58.92 & 84.30 & 15.51 & 32.90 \\
InstructBLIP 13B & 63.91 & 84.49 & 49.73 & 35.60 \\
\midrule
\textbf{Reproduction \& Optimization Procedure} \\
LLaVa v1.5 7B (Reproduction) & 52.95 & 86.57 & 60.87 & 54.43 \\
LLaVa v1.5 13B (Reproduction) & 65.38 & 86.94 & 64.13 & 56.81 \\
Single-Stage 7B & 51.47 & 86.57 & 61.63 & 54.85 \\
Single-Stage 13B & 70.05 & 87.00 & 63.26 & 57.25 \\
Frozen ViT (Single-Stage) & 51.47 & 86.57 & 61.63 & 54.85 \\
Finetune ViT (Multi-Stage) & 57.20 & 82.70 & 59.15 & 52.26 \\
Finetune ViT (Single-Stage) & 51.47 & 83.82 & 59.53 & 53.52 \\
\midrule
\textbf{Visual Representations} \\
IN1K ViT-L 224px & 51.47 & 82.08 & 52.95 & 50.01 \\
DINOv2 ViT-L 224px & 51.47 & 84.84 & 57.12 & 51.39 \\
CLIP ViT-L 224px & 51.47 & 85.80 & 59.09 & 53.90 \\
SigLIP ViT-SO 224px & 51.47 & 85.07 & 63.02 & 55.35 \\
\midrule
\textbf{Image Preprocessing} \\
CLIP ViT-L 336px (Letterbox) & 51.47 & 86.57 & 61.63 & 54.85 \\
CLIP ViT-L 336px (Resize Crop) & 51.47 & 85.42 & 61.24 & 53.52 \\
CLIP ViT-L 336px (Naive Resize) & 51.47 & 87.01 & 62.90 & 54.43 \\
SigLIP ViT-SO 384px (Letterbox) & 51.47 & 86.78 & 64.83 & 54.84 \\
SigLIP ViT-SO 384px (Resize Crop) & 51.47 & 84.62 & 62.94 & 54.51 \\
SigLIP ViT-SO 384px (Naive Resize) & 51.47 & 86.52 & 65.47 & 54.89 \\
\midrule
\textbf{Ensembling Visual Features} \\
CLIP 336px (Naive Resize) & 51.47 & 87.01 & 62.90 & 54.43 \\
DINOv2 + CLIP 336px (Letterbox) & 51.47 & 87.70 & 63.99 & 52.07 \\
DINOv2 + CLIP 336px (Naive Resize) & 51.47 & 87.29 & 65.02 & 52.52 \\
SigLIP 384px (Naive Resize) & 51.47 & 86.52 & 65.47 & 54.89 \\
DINOv2 + SigLIP 384px (Letterbox) & 51.47 & 87.89 & 67.19 & 55.43 \\
DINOv2 + SigLIP 384px (Naive Resize) & 51.55 & 88.30 & 67.63 & 54.82 \\
\midrule
\textbf{Base vs. Instruct Tuned LMs} \\
Vicuña v1.5 7B & 51.47 & 86.57 & 61.63 & 54.85 \\
Vicuña v1.5 13B & 70.05 & 87.00 & 63.26 & 57.25 \\
Llama-2 7B & 63.67 & 86.74 & 59.22 & 55.31 \\
Llama-2 13B & 65.71 & 86.91 & 62.54 & 56.51 \\
\midrule
\textbf{Better LMs} \\
Mistral v1 7B & 58.50 & 87.10 & 61.70 & 50.54 \\
Mistral Instruct v1 7B & 57.80 & 87.50 & 64.53 & 51.48 \\
\midrule
\textbf{Co-training on Language Safety Data} \\
Vicuña v1.5 7B & 51.47 & 86.57 & 61.63 & 54.85\\
Vicuña v1.5 7B (No Co-training) & 53.68 & 87.27 & 62.31 & 52.74 \\
Llama-2 7B & 63.67 & 86.74 & 59.22 & 55.31 \\
Llama-2 7B (No Co-training) & 67.18 & 86.88 & 57.17 & 53.87 \\
\midrule
\textbf{Scaling Training Time} \\
1 Epoch & 51.47 & 86.57 & 61.63 & 54.85 \\
1.25 Epochs & 51.80 & 86.80 & 61.69 & 54.02 \\
1.5 Epochs & 51.55 & 87.78 & 61.67 & 53.74 \\
2 Epochs & 53.93 & 87.03 & 62.52 & 54.91 \\
3 Epochs & 54.99 & 86.86 & 62.30 & 52.52 \\
\midrule
\textbf{Scaling Data} \\
Base & 51.47 & 86.57 & 61.63 & 54.85 \\
Base + LRV & 64.08 & 86.84 & 65.54 & 53.82 \\
Base + LVIS-4V & 51.47 & 86.98 & 62.60 & 53.40 \\
Base + LVIS-4V + LRV & 54.91 & 87.27 & 63.74 & 54.87 \\
\midrule
\textbf{Prism 7B} \\
Prism-CLIP 7B (Controlled) & 66.61 & 86.83 & 60.86 & 55.46 \\
Prism-CLIP 7B & 57.77 & 87.30 & 66.00 & 52.89 \\
Prism-SigLIP 7B (Controlled) & 65.14 & 87.07 & 64.54 & 55.48 \\
Prism-SigLIP 7B & 56.79 & 87.30 & 66.46 & 54.38 \\
Prism-DINOSigLIP 7B (Controlled) & 66.28 & 88.28 & 65.07 & 55.51 \\
Prism-DINOSigLIP 7B & 59.57 & 88.12 & 66.70 & 55.65 \\
\midrule
\textbf{Prism 13B} \\
Prism-CLIP 13B (Controlled) & 65.96 & 86.96 & 65.71 & 56.64 \\
Prism-CLIP 13B & 71.85 & 87.23 & 69.37 & 55.82 \\
Prism-SigLIP 13B (Controlled) & 62.85 & 86.82 & 62.90 & 55.48 \\
Prism-SigLIP 13B & 64.57 & 87.50 & 68.95 & 55.95 \\
Prism-DINOSigLIP 13B (Controlled) & 71.85 & 88.50 & 66.09 & 57.72 \\
Prism-DINOSigLIP 13B & 72.18 & 88.07 & 70.41 & 57.96 \\
    \bottomrule
  \end{tabular}
\label{tab:challenge-set-results}
\end{table}